\documentclass{article}

\usepackage[table,dvipsnames]{xcolor}

\usepackage{microtype}
\usepackage{graphicx}
\usepackage{subcaption}
\usepackage{booktabs}
\usepackage{tcolorbox}
\usepackage{multirow}
\usepackage{tikz}
\usepackage{array}
\usepackage{CJKutf8}

\tcbuselibrary{breakable}
\usetikzlibrary{arrows.meta,positioning,calc,backgrounds}

\usepackage{hyperref}

\usepackage[accepted]{icml2026}

\usepackage{amsmath}
\usepackage{amssymb}
\usepackage{mathtools}
\usepackage{amsthm}

\definecolor{rowBaseline}{HTML}{EEF1F5}
\definecolor{rowAblation}{HTML}{F4F2EE}
\definecolor{rowOurs}{HTML}{F2DFB6}

\usepackage[capitalize,noabbrev]{cleveref}

\theoremstyle{plain}
\newtheorem{theorem}{Theorem}[section]
\newtheorem{proposition}[theorem]{Proposition}

\theoremstyle{definition}
\newtheorem{definition}[theorem]{Definition}

\theoremstyle{remark}

\DeclareMathOperator{\Cov}{Cov}
\DeclareMathOperator{\gap}{Gap}
\newcommand{\kl}[2]{%
  \mathrm{KL}\!\left(#1\,\middle\|\,#2\right)%
}

\usepackage[textsize=tiny]{todonotes}

\icmltitlerunning{Rubric Curriculum RL: Exploiting the Generation Verification Gap in Non-Verifiable Domains}

\begin{document}

\twocolumn[
  % \icmltitle{\emph{Rubric Curriculum RL}: \\ Exploiting the Generation-Verification Gap in Creative Writing}
    % \icmltitle{\emph{Hillclimb Anything}: A Simple Method to Self-Improve on \\ Non-Verifiable Domains via the Generation-Verification Gap}
    % \icmltitle{\emph{Rubric Curriculum RL}: Self-Improvement Beyond Verifiable \\ Rewards via the Generation-Verification Gap}
    \icmltitle{\emph{Rubric Curriculum RL}: Exploiting the Generation-Verification \\ Gap in Non-Verifiable Domains}
    % Data-Free Self-Improvement on Non-Verifiable Domains via the Generation-Verification Gap
    % Exploiting the Generation-Verification \\ Gap for Data-Free Self-Improvement}
  % It is OKAY to include author information, even for blind submissions: the
  % style file will automatically remove it for you unless you've provided
  % the [accepted] option to the icml2026 package.

  % List of affiliations: The first argument should be a (short) identifier you
  % will use later to specify author affiliations Academic affiliations
  % should list Department, University, City, Region, Country Industry
  % affiliations should list Company, City, Region, Country

  % You can specify symbols, otherwise they are numbered in order. Ideally, you
  % should not use this facility. Affiliations will be numbered in order of
  % appearance and this is the preferred way.
\icmlsetsymbol{equal}{*}
\icmlsetsymbol{equaladvising}{\textdagger}
\begin{icmlauthorlist}
\icmlauthor{Sumeet Ramesh Motwani}{equal,oxf,msft}
\icmlauthor{Tejas Krishnan}{equal,oxf}
\icmlauthor{Alistair Letcher}{oxf}
\icmlauthor{Charles London}{oxf} \\
\icmlauthor{Suhaas Bhat}{oxf}
\icmlauthor{Huitian Jiao}{msft}
\icmlauthor{Philip Torr}{oxf}
\icmlauthor{Riashat Islam}{msft}
\icmlauthor{Christopher Summerfield}{oxf} \\
\icmlauthor{Jakob Foerster}{equaladvising,oxf}
\icmlauthor{Christian Schroeder de Witt}{equaladvising,oxf}
\icmlauthor{Qilong Gu}{equaladvising,msft}
\icmlauthor{Shital Shah}{equaladvising,msft}
\end{icmlauthorlist}

\icmlaffiliation{oxf}{University of Oxford}
\icmlaffiliation{msft}{Microsoft Research}
\icmlcorrespondingauthor{Sumeet Ramesh Motwani}{sumeet.motwani@eng.ox.ac.uk}
\icmlcorrespondingauthor{Tejas Krishnan}{tejas.krishnan81@gmail.com}

\icmlkeywords{reinforcement learning, post-training, language models, unverifiable domains, generation-verification gap, curriculum learning}

\vskip 0.3in
]

\printAffiliationsAndNotice{\icmlEqualContribution \textsuperscript{\textdagger}Equal advising. }

\begin{abstract}

Reinforcement learning with verifiable rewards (RLVR) has led to significant gains in math and code generation. Extending these gains to open-ended domains remains challenging: ground-truth verification is unavailable, human annotation is expensive, and learned reward models are prone to reward hacking. The generation-verification gap, where judging good outputs is easier than producing them, offers a potential path to improvement. However, rubric-based judges are often imprecise and stationary, so policies can overfit or reward hack them. We introduce Rubric Curriculum RL (RcRL), which exploits this gap without new data, human annotations, or stronger teacher models. We combine pairwise preferences over rollouts with a curriculum over rubric criteria, yielding a more discriminative signal than absolute scoring and a non-stationary objective. Whereas baselines plateau within a few hundred steps, RcRL preserves output entropy and keeps improving past 1000 steps. On creative writing, RcRL achieves a 70.5\% human win rate against the base model, with gains across benchmarks and judges. On HealthBench, RcRL improves over the base model by 14.6\% (full set) and 25.4\% (consensus subset), outperforming all training baselines, including instance-specific rubrics.

\end{abstract}

\section{Introduction}
\begin{figure}[t]
  \centering
    \includegraphics[width=1.0\columnwidth]{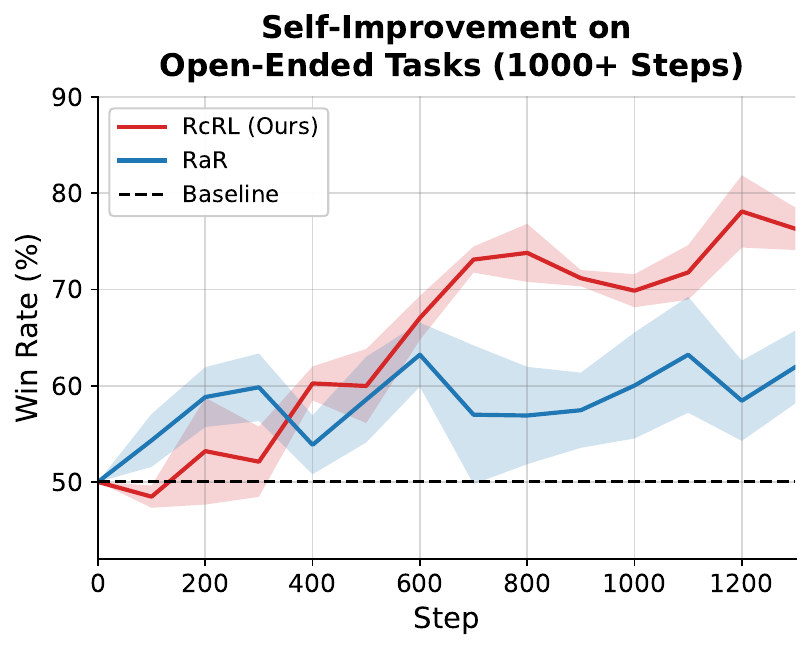}
    \caption{\textbf{RcRL vs. RaR on FictionBench.} Win rates against the base Qwen3-4B model over training, averaged across 5 seeds; shading indicates $\pm 1$ SEM. RcRL (red) improves over time, reaching $\sim$75\% by step 1300, while RaR (blue) plateaus early around 55--65\%. This demonstrates that curriculum learning over evaluation criteria enables \textbf{sustained improvements by allowing additional training compute to be spent productively.}}

  \label{fig:intro}
  \vspace{-10pt}
\end{figure}

\begin{figure*}[t]
  \centering
  \includegraphics[width=0.86\textwidth]{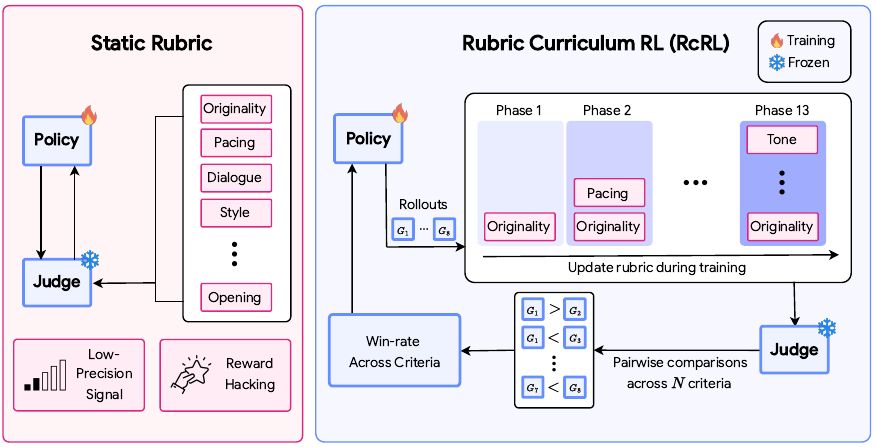}
  \caption{\textbf{Overview of Rubric Curriculum RL (RcRL).} (Left) The challenge: training with a static rubric leads to reward hacking, where the model optimizes for the reward signal without improving true quality. (Right) Our solution: RcRL progressively introduces evaluation criteria, using pairwise comparisons to allocate rewards. The curriculum advances when current criteria have improved (Section \ref{subsec:curriculum_pacing}).}
  \label{fig:method}
  \vspace{-0.4em}
\end{figure*}

Post-training language models with reinforcement learning has driven rapid progress on tasks where correctness can be checked automatically, for e.g. mathematical reasoning against known solutions or code execution. But a large fraction of real LLM usage \citep[56.4\%]{chatterji2025people} involves open-ended generation, for example creative writing, advice, brainstorming, explanations, etc. These tasks have no ground truth which makes them hard to hill-climb. Current methods to improve model capabilities on these open-ended tasks use either human annotations (which are expensive to obtain, especially online), offline datasets (need to be very large scale), or learnt reward models (prone to reward hacking). As RL continues to scale on verifiable domains, the gap between model performance on these open-ended domains and closed domains has widened \citep{zhang2026arenarl}, making it increasingly critical to bridge.

In this work, we focus on creative short-fiction writing and health question-answering as testbeds for open-ended self-improvement. These domains are among the most popular LLM applications, yet quality judgements are subjective and difficult to quantify, resulting in a noisy optimization landscape that is difficult to climb. The same problem arises in long-horizon open-ended tasks such as deep research and planning, where rubric rewards are increasingly used for training \citep{shao2026drtulureinforcementlearning, goel2025trainingaicoscientistsusing}. We discuss how RcRL can extend to these settings in Section \ref{sec:futurework}.

Our goal is to study how RL post-training can be used to improve a model’s open-ended generation capabilities using only self-generated data, with no human feedback or stronger teacher models.

Our approach exploits the generation-verification (GV) gap: for e.g. in creative writing, recognizing a compelling story is easier than writing one, even for the same model (see Section \ref{sec:gv_gap}). A model may reliably identify which of two stories is better yet consistently fail to generate stories at the level of the superior one. This asymmetry arises because verification requires compressing high-dimensional outputs into low-dimensional quality judgements to essentially rank outputs based on heuristics, while generation requires navigating a large search space to produce outputs that are both compelling and original. We formalize this gap and test it empirically in Sections \ref{sec:gv_gap} and \ref{sec:empirical}. Additionally, we provide a computational and cognitive science perspective of the generation-verification gap in Appendix \ref{app:cogsci}. 

This suggests that it might be possible to use a model’s own verification ability to supervise (and train) its generations. However, naive approaches to this fail because training a model with itself or a fixed reward model as a judge leads to reward hacking \citep{taylor2025schoolrewardhackshacking}. In open-ended domains, it is easy to find shortcuts that score well while producing degenerate outputs. Recent work has focused on training LLM Judges \citep{fein2025litbenchbenchmarkdatasetreliable} or using larger models with rubrics as judges \citep{gunjal2025rubricsrewardsreinforcementlearning} which provide richer signal during training, but gains still plateau within a few hundred steps of training with these methods.

We present Rubric Curriculum RL (RcRL), which addresses these failures through two mechanisms. First, we use the same model as a judge to provide pairwise preferences rather than absolute scores. In GRPO \citep{shao2024deepseekmath}, judging which of two rollouts is better is easier than assigning absolute scores and provides signal even when absolute quality is ambiguous (see Appendix \ref{app:pairwise-theory}). Next, we apply a curriculum over rubric criteria by introducing new criteria progressively. When a model begins overfitting to early criteria in the rubric, a new one is added which creates a non-stationary objective that is harder to reward hack \citep{gao2023scaling}. Rubrics have two primary issues, (i) they do not provide high precision rewards since judgments from multiple criteria lead to a single number causing a credit assignment problem and (ii) they act as stationary targets that can get reward hacked within a few hundred steps. RcRL follows a criteria weighting strategy that leads to more emphasis on the new criteria added as part of the reward signal, which allows the model to pick up on this criterion’s reward signal within a few steps of exploration.

This allows RcRL to scale RL training far beyond prior methods. Standard self-improvement approaches plateau or collapse within 50–250 steps; RcRL continues improving past 1000 steps while preserving output entropy. In human evaluations, RcRL-trained models achieve a 70.5\% win rate over the base instruction-tuned model. Across creative writing benchmarks and HealthBench \citep{arora2025healthbench}, RcRL outperforms other strong absolute reward or non-curriculum based baselines by a large and consistent margin.

Our contributions are as follows:
\begin{enumerate}
\item We characterize and empirically test the \emph{generation-verification gap} in open-ended domains.
\item We introduce RcRL, which combines pairwise rewards with a curriculum over rubric criteria. The curriculum leads to high precision signal that reduces reward hacking during training.
\item We demonstrate that RcRL sustains improvement over 1000+ training steps, preserves output entropy, outperforms self-improvement baselines across three creative writing benchmarks, and achieves a 70.5\% human-evaluated win rate—all \emph{without new data, human annotations, or stronger judge models}.
\item We extend RcRL to HealthBench and show that it consistently outperforms strong baselines.
\end{enumerate}

\section{Related Work}
\paragraph{RL on Open-Ended Domains.}
RLVR has driven substantial progress on verifiable tasks like mathematics and code \citep{guo2025deepseek}, but extending RL to open-ended domains remains difficult. RLHF can improve subjective quality \citep{ziegler2020finetuninglanguagemodelshuman, bai2022traininghelpfulharmlessassistant}, but human annotation is expensive offline and impractical for online RL. Alternatives include LLM-as-judge approaches \citep{zheng2023judgingllmasajudgemtbenchchatbot, lee2024rlaifvsrlhfscaling}, trained reward models \citep{fein2025litbenchbenchmarkdatasetreliable, liao2025rlmrreinforcementlearningmixed, wu2025longwriterzeromasteringultralongtext}, and self-rewarding methods for instruction tuning \citep{yuan2025selfrewardinglanguagemodels}. However, these approaches either require stronger models, additional training data, or produce unreliable signal. Part of the difficulty is that creativity and open-ended performance itself resists a precise definition beyond novelty and usefulness \citep{schmidhuber2010formal}. Recent work has explored equipping judges with rubrics to provide more structured feedback \citep{gunjal2025rubricsrewardsreinforcementlearning}. We build on this direction, introducing pairwise evaluation and a curriculum over rubric criteria that enables sustained training without reward hacking.

\paragraph{Generation-Verification Gap and Reward Hacking.}
Generation is widely believed to be harder than verification; if $P \neq NP$, such problems must exist \citep{hartmanis1993godel, cook2023complexity}. From a learning-theoretic perspective, verifiers require lower-capacity function classes and thus fewer samples to learn \citep{vapnik2015uniform}. This gap appears during fine-tuning \citep{swamy2025all} and has been exploited for self-improvement in verifiable domains \citep{song2025mindgapexaminingselfimprovement}. The gap also exists in open-ended domains \citep{baer2014gold}, but exploiting it is harder because reward hacking emerges: models find shortcuts that increase the proxy reward without improving true quality. \citet{gao2023scaling} derive scaling laws showing that proxy rewards rise while true quality degrades. \citet{skalse2022defining} formalize when a proxy is hackable relative to a true reward, and \citet{rafailov2024scaling} show that even direct alignment methods such as DPO exhibit similar degradation at high KL budgets. This failure mode has been documented specifically for LLM judges in open-ended settings \citep{pan2024spontaneousrewardhackingiterative, wolf2025rewardmodeloveroptimisationiterated, zhao2025tokenfoolllmasajudge}. Our curriculum addresses this with a non-stationary objective that prevents the policy from over-optimizing any single fixed criterion.

\paragraph{Curriculum Learning and Exploration.}
Curriculum learning (training on progressively harder examples)  \citet{bengio2009curriculum} shows that ordering data by difficulty improves convergence and generalization; \citet{soviany2022curriculum} survey applications across deep learning. In RL, curriculum methods have been applied to goal generation \citep{florensa2017reverse}, environment complexity \citep{openai2019solving}, and task difficulty \citep{narvekar2020curriculum}. Recent RLVR work shows that curriculum improves reasoning \citep{parashar2025curriculumreinforcementlearningeasy} and matches dense-reward sample efficiency \citep{h1paper}. A separate challenge is that RL training tends to reduce output diversity \citep{kirk2024understanding, yu2025dapo}; in GRPO specifically, entropy collapse occurs as the policy becomes overly deterministic. Entropy regularization can counteract this \citep{ahmed2019understanding, simoni2025gtpostabilizinggrouprelative, tuyls2025representationbasedexplorationlanguagemodels}, and prior work suggests that curriculum structure itself helps preserve entropy \citep{bengio2009curriculum, cao2026efficientreinforcementlearningsemantic, setlur2025e3learningexploreenables}. Our approach combines both insights: the curriculum over rubric criteria repeatedly shifts the reward function \citep{zhou2022continuouslydiscoveringnovelstrategies}, slowing convergence to exploitative strategies while maintaining output diversity.

\section{Preliminaries and Method}
\label{sec:gv_gap}
\subsection{Understanding the Generation-Verification Gap}
For many domains, verification is considered easier than generation. We hypothesize that a similar asymmetry exists in creative writing. One way to understand this gap is that generation involves searching over a vast space of possible answers, whereas evaluation does not (see Appendix \ref{app:cogsci}). This asymmetry increases in creative writing, where the goal is to be both useful/good and novel \citep{schmidhuber2010formal}, where the latter inherently involves search. To make this intuition precise and empirically testable, we formalize the \textit{generation-verification gap}:

Let $\mathcal{X}$ be a distribution over prompts, $\mathcal{Y}$ the space of responses, and $Q: \mathcal{X} \times \mathcal{Y} \to [0, 1]$ a quality function. In our experiments, $Q$ is instantiated as the normalized rubric score from a strong external judge (e.g., GPT-5.2). Let $\pi_\mathcal{M}(\cdot|x)$ denote the model's output distribution given prompt $x$.

\begin{definition}[Generative Ability]
The expected quality of a single generation from model $\mathcal{M}$:
\[\bar{G}_\mathcal{M} = \mathbb{E}_{x \sim \mathcal{X}}\left[\mathbb{E}_{y \sim \pi_\mathcal{M}(\cdot|x)}\left[Q(x, y)\right]\right]\]
\end{definition}

\begin{definition}[Self-Verification Ability]
Expected quality of response selected by $\mathcal{M}$ from $N$ independent draws:
\[\bar{V}_\mathcal{M}^{(N)} = \mathbb{E}[Q(x,y_{i^*})], \quad i^* \in \arg\max_{i} \mathrm{rank}_\mathcal{M}(x, y_i),\]
where $x \sim \mathcal{X}$, $y_1, \ldots, y_N \overset{\text{i.i.d.}}{\sim} \pi_\mathcal{M}(\cdot|x)$, and $\mathrm{rank}_\mathcal{M}$ denotes the model's ranking over responses via some self-selection method.
\end{definition}

\begin{definition}[Generation-Verification Gap]
A model $\mathcal{M}$ exhibits a generation-verification gap on quality function $Q$ if for some $N$,
\[\bar{V}_\mathcal{M}^{(N)} - \bar{G}_\mathcal{M} > 0.\]
That is, the model can reliably select responses of higher quality than it produces on a single draw.
\end{definition}

\begin{figure}[h]
  \centering
  \includegraphics[width=1.0\columnwidth]{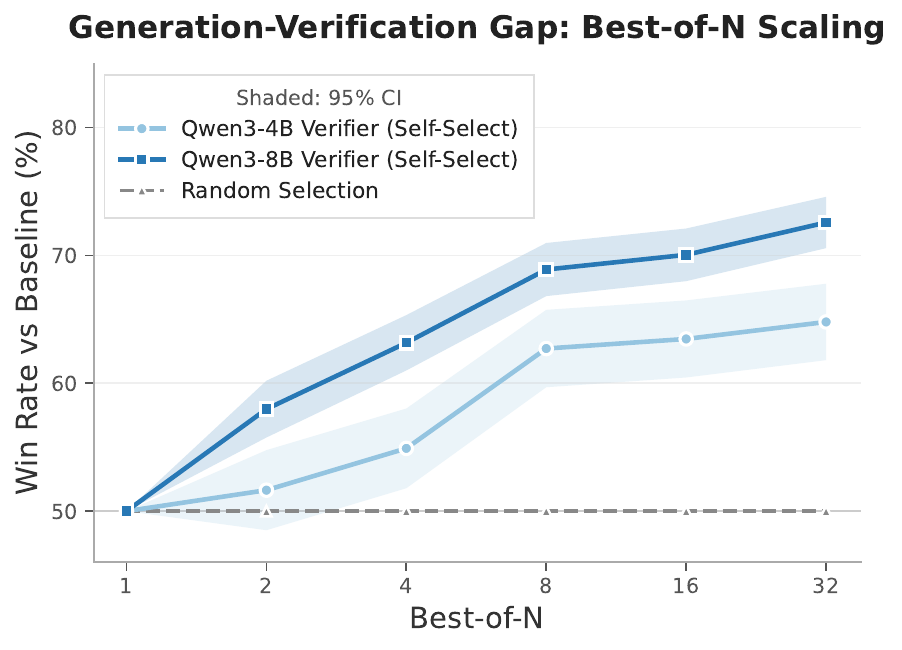}
  \caption{\textbf{Best-of-$N$ inference scaling on FictionBench.} Win rate of self-selected Best-of-$N$ against the $N=1$ baseline, judged by GPT-5.2 as ground truth. Win rates exceed 50\% from $N=2$ onwards (Claim 1), grow monotonically with $N$, and are uniformly larger for the 8B model than the 4B model (Claim 2).}
  \label{fig:inference-scaling}
\end{figure}

\paragraph{Experimental Framework:}
To empirically establish the \emph{generation-verification gap} in creative writing, we test the following claims:

\textbf{Claim 1 (Gap Exists):} The model can self-select responses of higher quality than it produces on a single draw:
\[\bar{V}_\mathcal{M}^{(N)} > \bar{G}_\mathcal{M}.\]
\textit{Evidence:} Best-of-$N$ improves over $N=1$ (see Figure \ref{fig:inference-scaling})

For a prompt $x$, we generate $N$ independent rollouts $\{y_1, y_2,  \dots, y_N\}$ from the base model $\pi_\mathcal{M}$ and use $\pi_\mathcal{M}$ itself as a verifier to select the ``best'' response. Selection is via pairwise comparisons between all rollouts ($\binom{N}{2}$ comparisons), where the rollout with the highest win rate is chosen. Finally, we measure the win rate of the selected $y_i$ against 5 held out evaluation stories for prompt $x$, as judged by the oracle (GPT-5.2).

\textbf{Claim 2 (Gap Scales):} The gap grows with \emph{model size} and with the \emph{number of samples $N$}:
\[\bar{V}_\mathcal{M}^{(N)} - \bar{G}_\mathcal{M} \;\text{ increases in both } N \text{ and model scale}.\]
\textit{Evidence:} In Figure \ref{fig:inference-scaling}, Qwen3-8B exhibits a larger gap than Qwen3-4B at every $N$, and both curves are monotonically increasing in $N$.

\textbf{Claim 3 (Gap Is Exploitable):} Using self-verification as a reward signal improves creative writing capabilities:
\[\bar{G}_{\mathcal{M}'} > \bar{G}_{\mathcal{M}} \quad \text{after RL training}.\]
\textit{Evidence:} RcRL achieves a 70.5\% win rate against Qwen3-4B on human evaluations, with significant improvements across all tested benchmarks (Section \ref{sec:empirical}).

\subsection{RL on Qualitative Domains}

To exploit the generation-verification gap, we employ a self-improvement framework where the model provides relative judgments over a group of its own outputs. We use Dr. GRPO \citep{liu2025understandingr1zeroliketrainingcritical} which is well-suited to this setup as it directly leverages within-group variance in output quality. For a prompt $\mathbf{q}$ and a group of $G$ rollouts $\{o_1, \ldots, o_G\}$, the RL training objective is:
$$\frac{1}{G} \sum_{i=1}^{G} \sum_{t=1}^{|\mathbf{o}_i|} \min \left[ \frac{\pi_\theta(o_{i,t} | \mathbf{q}, \mathbf{o}_{i,<t})}{\pi_{\theta_{old}}(o_{i,t} | \mathbf{q}, \mathbf{o}_{i,<t})} \hat{A}_{i,t}, \; \tilde{\rho}_{i,t} \hat{A}_{i,t} \right]$$

where $\tilde{\rho}_{i,t}$ is the probability ratio clipped to $[1-\epsilon, 1+\epsilon]$,
$$\hat{A}_{i,t} = R(\mathbf{q}, \mathbf{o}_i) - \text{mean}\left(\{R(\mathbf{q}, \mathbf{o}_1), \ldots, R(\mathbf{q}, \mathbf{o}_G)\}\right)$$
Under the Rubrics as Rewards (RaR; \citep{gunjal2025rubricsrewardsreinforcementlearning}) framework, rewards at each step are provided as follows:
$$
R_{\mathrm{RaR}}(\mathbf{q}, \mathbf{o}_i; \mathbf{r})
\;:=\;
\pi_{\mathrm{judge}}\!\left(\mathbf{q}, \mathbf{o}_i, \mathbf{r}\right)
\in [0,1].
$$
Here, $\mathbf{r}$ denotes the rubric provided to the judge (i.e., $\pi_{\mathrm{judge}}$ is conditioned on the question $\mathbf{q}$, the rollout $\mathbf{o}_i$, and the rubric $\mathbf{r}$). A rubric can be seen as a checklist of criteria that specifies what counts as a high quality output. In RaR, $\pi_{\mathrm{judge}}$ is often instantiated as a separate, larger and stronger model than the policy being trained (i.e., $\pi_{\mathrm{judge}} \neq \pi_\theta$). This assumes the existence of stronger models or more data.

\begin{figure}[h]
  \centering
  \includegraphics[width=1.0\columnwidth]{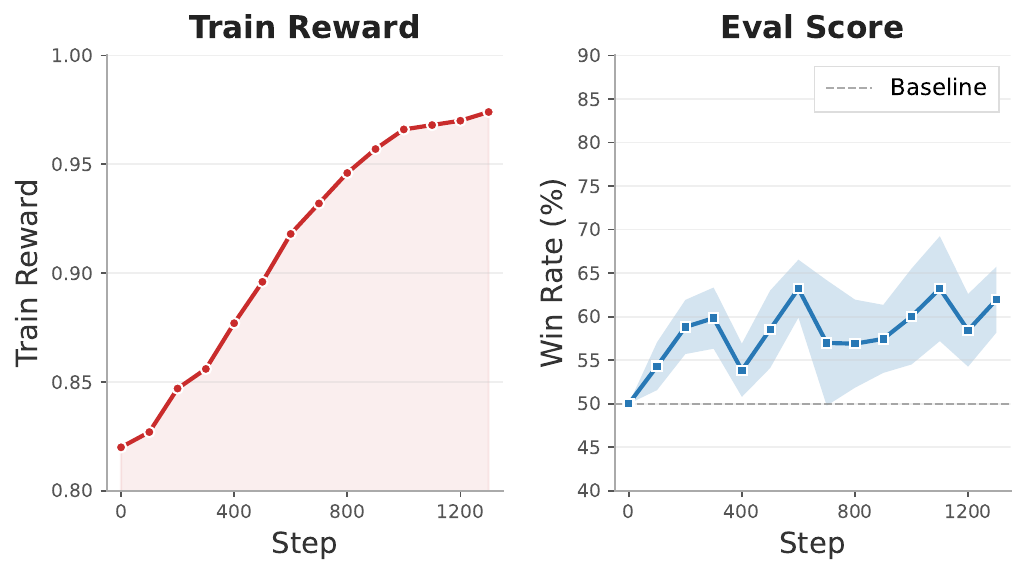}
  \caption{\textbf{Reward hacking with static rubrics.} (Left) Absolute train reward increases steadily over the course of training, reaching near-saturation. (Right) Evaluation win rate against the base model on FictionBench shows no consistent improvement as training progresses, indicating a performance mismatch. }
  \label{fig:RaR-reward-hacking}
\end{figure}

While RaR provides a clear training signal, it is prone to reward hacking in qualitative domains like creative writing. As illustrated in Figure \ref{fig:RaR-reward-hacking}, training reward climbs steadily while evaluation win rates remain volatile and saturate after a few hundred steps, demonstrating a critical decoupling between reward optimization and true quality improvement \citep{gao2023scaling}.
This decoupling has two sources. First, with a fixed rubric $\mathbf{r}$
and judge $\pi_{\text{judge}}$, the model shifts from learning to write better to exploiting specific biases in $\pi_{\text{judge}}$ \citep{singhal2023long, skalse2022defining}, leading to the entropy degradation observed in Section \ref{sec:empirical}. In Section \ref{sec:analysis} we study this performance degradation on a per-criterion level. Second, RaR relies on absolute rewards which present a stationary target that gets reward hacked  quicker than it would for pairwise rewards \citep{karwowski2023goodhartslawreinforcementlearning, zhao2025tokenfoolllmasajudge, fu2026rewardshapingmitigatereward}. Early work from \cite{xu2025unifiedpairwiseframeworkrlhf} has shown pairwise rewards as a promising approach to reduce reward hacking. Intuitively, relative rewards are more resistant to reward hacking because the comparison target evolves with the policy itself (see Appendix \ref{app:pairwise-theory}). Unlike absolute rewards which can reach a maximum threshold, pairwise rewards do not saturate in a similar way.

\subsection{Our method}
\label{sec:ourmethod}
Rewards in our method are produced by a fixed self-judge $\pi_{\theta_0}$ that compares two rollouts at a time. For a single rubric item,
a rollout's scalar reward is its empirical win rate against all the other rollouts. Importantly, instead of providing a noisy stationary optimization target with the entire rubric at once, we use a \emph{curriculum over rubric criteria}.
We let $\pi_{\mathrm{judge}} := \pi_{\theta_0}$ denote the fixed (initial) snapshot of the base model, held constant during training (i.e., not updated online). Let $\mathcal{R}=\{r^{(1)},r^{(2)},\ldots\}$ be a sequence of rubric items, and let $k(t)$ denote the active rubric index at training step $t$ (updated every $n$ steps).
Given a set of $G$ rollouts $\{\mathbf{o}_1,\ldots,\mathbf{o}_G\}$ generated by $\pi_\theta$ for the same query $\mathbf{q}$, let $y^{(m)}_{i \succ j} \in \{0,1\}$ indicate whether $\pi_{\mathrm{judge}}$ prefers $\mathbf{o}_i$ over $\mathbf{o}_j$ under rubric item $r^{(m)}$. The win-rate reward under that item is the average preference against the other rollouts: 
\[
R^{(m)}_{\mathrm{pair}}(\mathbf{q}, \mathbf{o}_i)
=
\frac{1}{G-1}
\sum_{\substack{j=1 \\ j \neq i}}^{G}
y^{(m)}_{i \succ j}.
\]
To improve the signal on new rubric items and prevent degradation on the previous criteria, we use a weighted curriculum over rubric items: at step $t$, the newest rubric item $r^{(k(t))}$ receives half of the total weight, while the remaining half is distributed uniformly over all previous items $\{r^{(1)},\ldots,r^{(k(t)-1)}\}$. For \(k(t)=1\), we set \(w^{(1)}_t=1\). For \(k(t)>1\), we use weights
$$
w^{(m)}_t
:=
\begin{cases}
\frac{1}{2}, & m = k(t),\\[4pt]
\frac{1}{2\,(k(t)-1)}, & 1 \le m < k(t),
\end{cases}
\;\;
\textstyle\sum_{m=1}^{k(t)} w^{(m)}_t = 1
$$
The resulting weighted pairwise curriculum reward is:
$$
R_{\mathrm{W\text{-}pair}}(\mathbf{q}, \mathbf{o}_i; t)
\;:=\;
\sum_{m=1}^{k(t)} w^{(m)}_t \, R^{(m)}_{\mathrm{pair}}(\mathbf{q}, \mathbf{o}_i)
\;\in\;[0,1].
$$
We present the algorithm for RcRL below. The curriculum advances every $n$ training steps, introducing a new criterion while re-weighting all previous ones. See Appendix \ref{app:theory} for further analysis on the benefits of a curriculum. We set $n=100$ to allocate sufficient training compute without over-optimizing on a single criterion (see Section \ref{sec:analysis}).

\begin{algorithm}[h]
  \caption{Rubric Curriculum RL (RcRL)}
  \label{alg:rcrl}
  \begin{algorithmic}[1]
    \STATE \textbf{Input:} Policy $\pi_\theta$, rubric $\mathcal{R} = \{r^{(1)}, \ldots, r^{(K)}\}$, prompt distribution $\mathcal{X}$
    \STATE \textbf{Init:} $k(t) \leftarrow 1$, judge $\pi_{\mathrm{judge}} \leftarrow \pi_{\theta_0}$ (frozen)
    \FOR{step $t = 1$ \textbf{to} $T$}
      \STATE Sample $\mathbf{q} \sim \mathcal{X}$; generate $\{\mathbf{o}_1, \ldots, \mathbf{o}_G\} \sim \pi_\theta(\cdot \mid \mathbf{q})$
      \STATE Compute $R_{\mathrm{W\text{-}pair}}(\mathbf{q}, \mathbf{o}_i; t)$ for all $i$
      \STATE Update $\theta$ via GRPO with rewards $\{R_{\mathrm{W\text{-}pair}}\}_{i=1}^G$
      \IF{$t \mod n = 0$ \textbf{and} $k(t) < K$}
        \STATE $k(t) \leftarrow k(t) + 1$ \hfill $\triangleright$ Expand curriculum
      \ENDIF
    \ENDFOR
    \STATE \textbf{Return:} $\pi_\theta$
  \end{algorithmic}
\end{algorithm}

\section{Empirical Results}
\label{sec:empirical}
\subsection{Evaluation Framework}
To empirically validate RcRL, we focus primarily on creative writing with a detailed set of experiments and ablations across three distinct benchmarks detailed below. In Section~\ref{sec:cross-domain}, we further evaluate cross-domain generalization on HealthBench, a health conversation benchmark.
\begin{itemize} 
    \item \textbf{Creative Writing V3 (CWv3):} A comprehensive suite \citep{paech2024creative} that first evaluates model responses against a multi-point rubric and subsequently computes ELO ratings. It provides three primary metrics: \textbf{rubric score}, \textbf{win rate}, and \textbf{ELO rating}. To ensure results are robust to specific model biases, we conduct evaluations using three separate frontier model judges.
    \item \textbf{FictionBench:} An independently developed benchmark using pairwise comparisons, which better capture qualitative differences than absolute scores in creative domains \citep{zhang2026arenarl}. Each evaluation randomly samples 20 prompts from a pool, generating two samples per model per prompt; GPT-5.2 evaluates all 40 pairs using 7$\times$ oversampling (280 total comparisons). We report \textbf{win rates} and \textbf{ELO ratings} via Bradley-Terry \citep{bradley1952rank}. As an independently constructed prompt set, FictionBench also mitigates benchmark leakage risks.
    \item \textbf{LechMazur (LM):} A public benchmark \citep{mazur2024lech} focused on high-dimensional constraints in writing. Using GPT-5.2 as a judge, it evaluates model performance across 18 distinct criteria, producing both an aggregate \textbf{rubric score} and a fine-grained \textbf{heatmap}. The first 8 criteria measure major story aspects while the latter 10 measure element integration from the prompt.
\end{itemize}

We conduct our primary experiments and analysis on Qwen3-4B. To evaluate scaling behavior, we train models across the Qwen3 family \citep{{yang2025qwen3technicalreport}} ranging from 0.6B to 14B parameters, and additionally validate cross-architecture generalization on the Llama family. Note, all baseline models come instruction-tuned.

\subsection{Main Creative Writing Results}

RcRL substantially outperforms both the base instruction-tuned Qwen3-4B and the RaR baseline across all evaluation settings. On CWv3 (Table \ref{tab:main-results}),  RcRL achieves consistent gains across all 3 independent judges, with ELO improvements of +577 (GPT-5.2), +105 (Claude 4.5 Opus), and +524 (Deepseek V3.2) over the base model. Notably, RcRL's ELO gains are 1.6–2.7$\times$ larger than RaR's, with win rates consistently favoring RcRL.

\begin{table}[t]
  \caption{\textbf{Additional benchmark results.} FictionBench and the LechMazur Writing benchmark. Best results per model in \textbf{bold}.}
  \begin{center}
    \begin{small}
      \begin{sc}
        \resizebox{\columnwidth}{!}{%
        \begin{tabular}{l cc c c}
          \toprule
          & \multicolumn{2}{c}{\textbf{FictionBench}} & \textbf{LechMazur} & \\
          \cmidrule(lr){2-3} \cmidrule(lr){4-4}
          Method & ELO & Win\% & Score & $\mathcal{H}_{\text{ratio}}$ \\
          \midrule
          Qwen 3 4B (Baseline)      & 1000 & --   & 77.0 & -- \\
          Qwen 3 4B RaR  & 1052 & 57.5 & 80.3 & 0.73 \\
          \rowcolor{red!7} \textbf{Ours (RcRL)}    & \textbf{1181} & \textbf{73.9} & \textbf{80.9} & \textbf{0.93} \\
          \bottomrule
        \end{tabular}%
        }
      \end{sc}
    \end{small}
  \end{center}
\label{tab:additional-benchmarks}
\end{table}

On FictionBench (Table \ref{tab:additional-benchmarks}), RcRL achieves a 73.9\% win rate against the base model, with a 3.5$\times$ ELO gain over RaR. On LechMazur, RcRL scores 80.9 compared to 77.0 for the base model, with particularly strong gains in prompt element integration (Q9A–Q9J; see Appendix \ref{app:lechmazur} for the full heatmap). Section \ref{sec:analysis} presents ablations isolating each component and evaluates additional baselines. See Appendix \ref{app:repetition} for lexical analysis of trained models' outputs.

\begin{table*}[t]
  \caption{\textbf{Creative Writing Bench v3 evaluation results.} Three frontier models (GPT-5.2, Claude 4.5 Opus, Deepseek V3.2) serve as independent judges, evaluating outputs from each method via pairwise comparisons. We report ELO ratings (baseline anchored at 1000), rubric scores (0--100 scale), and win rates against the Qwen 3 4B baseline (\%). ELO ratings reflect relative rankings within each judge and are not comparable across judges. Best results per metric and judge in \textbf{bold}.}
  \begin{center}
    \begin{small}
      \begin{sc}
        \resizebox{\textwidth}{!}{%
        \begin{tabular}{l rrr rrr rrr}
            \toprule
            & \multicolumn{3}{c}{GPT-5.2} & \multicolumn{3}{c}{Claude 4.5 Opus} & \multicolumn{3}{c}{Deepseek V3.2} \\
            \cmidrule(lr){2-4} \cmidrule(lr){5-7} \cmidrule(lr){8-10}
            Method & ELO & Rubric & Win\% & ELO & Rubric & Win\% & ELO & Rubric & Win\% \\
            \midrule
            Qwen 3 4B (Baseline)    & 1000 & 45.4 & -- & 1000 & 40.8 & -- & 1000 & 51.2 & -- \\
            Qwen 3 4B RaR & 1210 & 48.5 & 61.1 & 1051   & \textbf{41.1} & 51.7 & 1326   & 55.0 & 53.9 \\
            \rowcolor{red!7} \textbf{Ours (RcRL)}   & \textbf{1577} & \textbf{50.1} & \textbf{69.3} & \textbf{1105} & 40.0 & \textbf{59.2} & \textbf{1524} & \textbf{58.1} & \textbf{65.4} \\
                        \bottomrule
        \end{tabular}%
        }
      \end{sc}
    \end{small}
  \end{center}
\label{tab:main-results}

\end{table*}

\subsection{Pairwise Model Comparisons}
\begin{figure}[h]
  \centering
  \includegraphics[width=1.0\columnwidth]{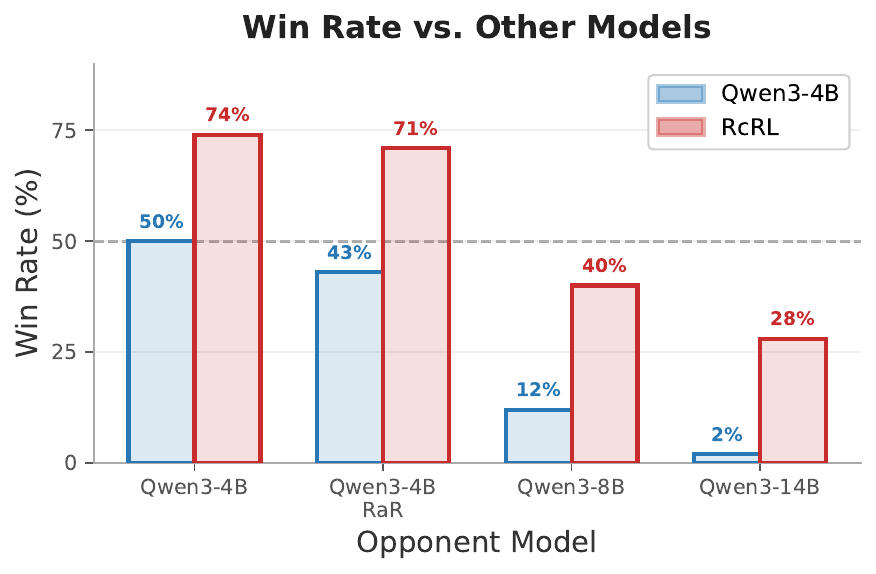}
  \caption{\textbf{Head-to-head win rates on FictionBench.} RcRL achieves substantial gains over the Qwen3-4B baseline against all opponents: +28\% against RaR, +28\% against Qwen3-8B, and +26\% against Qwen3-14B.}
  \label{fig:win_rates}
  % \vspace{-10pt}
\end{figure}

Beyond aggregate benchmark scores, we evaluate RcRL through direct pairwise comparisons against models from the Qwen3 family. Figure \ref{fig:win_rates} presents win rates on FictionBench. RcRL achieves a 71.4\% win rate against the RaR baseline. More important is RcRL’s performance against larger models: while Qwen3-4B wins only 12.1\% and 1.8\% of the time against 8B and 14B respectively, RcRL raises these to 40.0\% and 27.9\%, substantially closing the model capacity gap without new data or teacher models.

\begin{figure}[h]
  \centering
  \includegraphics[width=1.0\columnwidth]{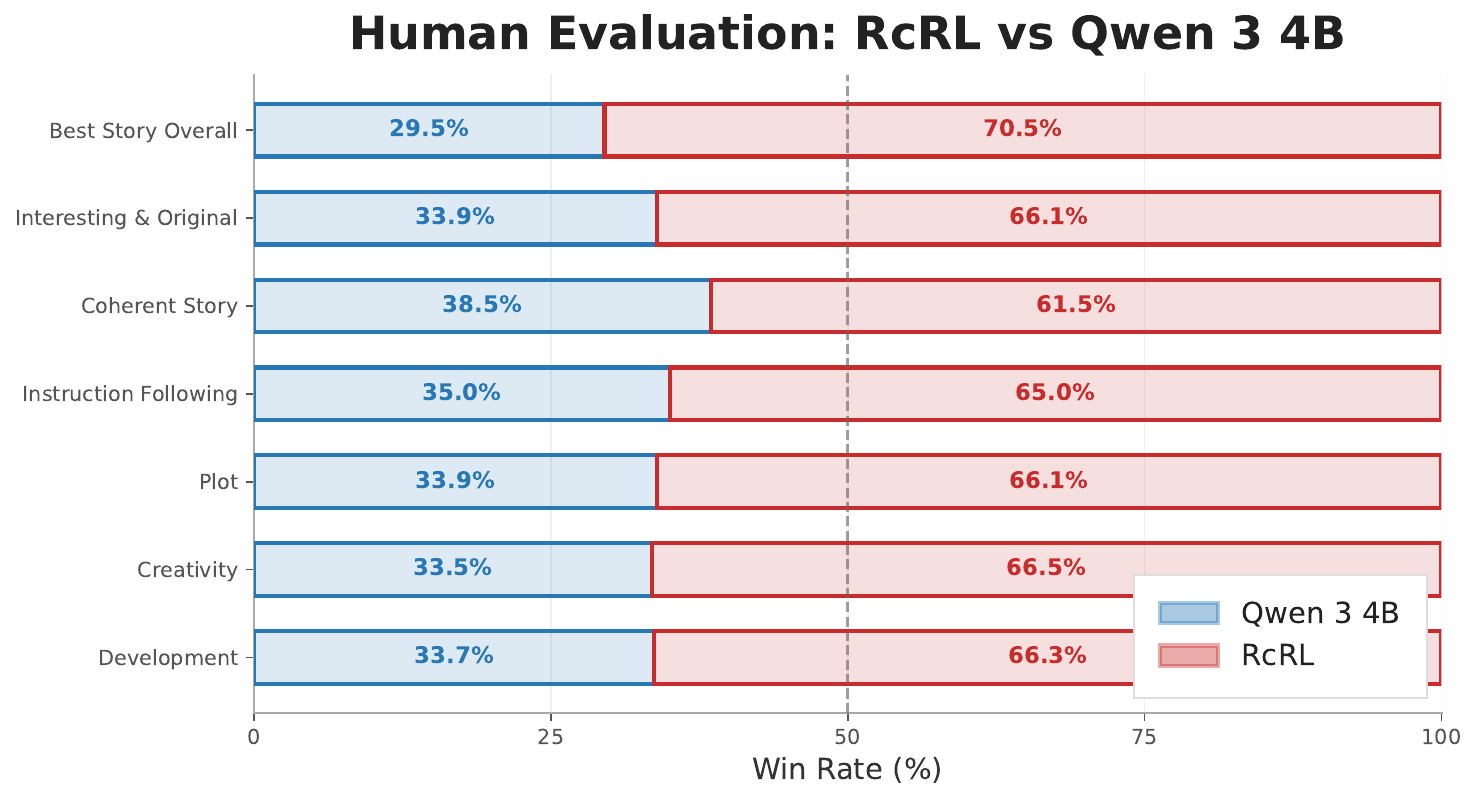}
  \caption{\textbf{Human evaluation results.} Win rates of RcRL-trained Qwen3-4B against the base model across seven criteria, averaged over three annotators who conducted 500 comparisons each. RcRL achieves a \textbf{70.5\% win rate on \textit{Best Story Overall}}, with consistent improvements across all criteria.}
  \label{fig:human-eval-results}
  \vspace{-10pt}
\end{figure}

\paragraph{Human Evaluation.} To validate that automated judgments reflect genuine quality improvements aligned with human preferences, we conducted a human evaluation study. Three professional annotators evaluated 500 distinct prompt-response pairs each, comparing RcRL outputs against Qwen3-4B baseline outputs. For each pair, annotators rated which response was superior across seven criteria, including an overall preference judgment. Presentation order was randomized to control for position bias. 
Human judges preferred RcRL outputs \textbf{70.5\%} of the time overall (Figure \ref{fig:human-eval-results}), with unanimous agreement among all three annotators on 70.8\% of comparisons. Table \ref{tab:human-eval} in Appendix \ref{app:human-eval} provides the full per-criterion breakdown and statistics.
\subsection{Scaling Behavior}
\begin{figure}[h]
  \centering
  \vspace{-0.14em}
  \includegraphics[width=0.95\columnwidth]{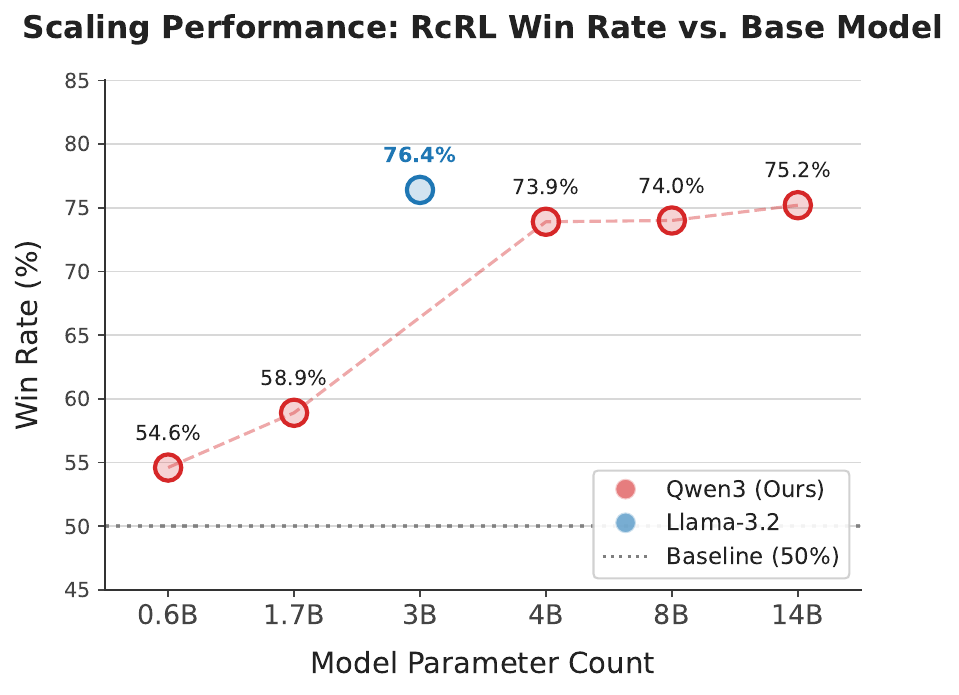}
  \caption{\textbf{Scaling behavior of RcRL.} Win rates against base models across model sizes for Qwen3 (red) and Llama-3.2 (blue). Performance improves with scale, from 54.6\% at 0.6B to 75.2\% at 14B parameters. The Llama-3.2-3B result (76.4\%) demonstrates that RcRL generalizes across model families.}
  \label{fig:gv-gap-scaling}
  \vspace{-0.6em}
\end{figure}

RcRL directly exploits the generation-verification gap by using verification (pairwise comparisons) to supervise generation. Previous work has shown the gap on verifiable domains scales with model size \citep{song2025mindgapexaminingselfimprovement}, so we expect RcRL to yield increasing benefits at scale. Figure \ref{fig:gv-gap-scaling} confirms this: RcRL’s win rate against base models increases monotonically from 54.6\% at 0.6B to 75.2\% at 14B parameters. Cross-architecture generalization is supported by Llama-3.2-3B showing similar improvements.

\section{Analysis}
We give the theoretical analysis how the generation-verification gap can lead to self-improvement in Appendix \ref{app:theory}, and of why pairwise rewards outperform absolute ones in Appendix \ref{app:pairwise-theory}. In this section, we test each component empirically through ablations, scaling experiments, and analyses of training dynamics.

\label{sec:analysis}

\subsection{Ablations}
Table \ref{tab:ablations} isolates the contribution of RcRL's core components. Removing pairwise rewards in Absolute RcRL (using absolute rubric scores instead) reduces FictionBench win rate by 18.2\%, confirming the advantage of relative comparisons. Removing the curriculum significantly reduces ELO ratings and causes rapid entropy collapse ($\mathcal{H}_{\text{ratio}} = 0.83$ within 300 steps), validating that progressive criteria addition is essential for preserving output diversity and sustaining gains. We also apply a Dropout baseline where we randomly drop criteria from RaR to make the objective non-stationary. Furthermore, we test 3 random curriculum orderings (post our main results) and show that performance remains strong across orderings (described in Appendix \ref{app:rubric-criteria}). On RaR, we perform an LR sweep and report results for the best LR. For RcRL, we use a default LR of 5e-6.

We use exploration bonuses \citep{tuyls2025representationbasedexplorationlanguagemodels} and usefulness rewards derived from early work from \cite{schmidhuber2010formal} as add-ons to our method and see that they sometimes offer strong gains, especially together. We describe the exact structure of these techniques in Appendix \ref{app:auxiliary-rewards}

\begin{table*}[t]
  \caption{\textbf{Ablation study and baseline comparisons.} We evaluate the contribution of each component of RcRL. ``$+$'' denotes adding a component to the base method, ``$-$'' denotes removing from the full method. All models trained from Qwen 3 4B. Win rates computed against the base model. Best results in \textbf{bold}.}
  \begin{center}
    \begin{small}
      \begin{sc}
      \renewcommand{\arraystretch}{0.9}
        \resizebox{0.9\textwidth}{!}{%
        \begin{tabular}{p{5.2cm} cc c<{\hspace{0.875em}} >{\hspace{0.875em}}c cc}
          \toprule
          & \multicolumn{3}{c}{\textbf{CW Bench v3}} & \multicolumn{3}{c}{\textbf{FictionBench + LM}} \\
          \cmidrule(lr){2-4} \cmidrule(lr){5-7}
          Method & ELO & Rubric & Win\% & ELO & Win\% & LM \\
          \midrule
          \rowcolor{gray!15} \multicolumn{7}{l}{\textit{No RL Training}} \\
          \midrule
          Qwen 3 4B (Baseline)   & 1000 & 45.4 & --   & 1000 & --   & 77.0 \\
          \midrule
          \rowcolor{gray!15} \multicolumn{7}{l}{\textit{No Curriculum}} \\
          \midrule
          Absolute RaR       & 1210 & 48.5 & 61.1 & 1052 & 57.5 & 80.3 \\
          Dropout Absolute RaR & 1166 & 49.8 & 55.4 & 1113 & 65.7 & 80.9 \\
          Pairwise RaR       & 1195 & 47.6 & 68.2 & 1144 & 69.6 & 82.0 \\
          \midrule
          \rowcolor{gray!15} \multicolumn{7}{l}{\textit{No Pairwise Rewards}} \\
          \midrule
          Absolute RcRL      & 1357 & 49.8 & 61.4 & 1040 & 55.7 & 79.6 \\
          \midrule
          \rowcolor{gray!15} \multicolumn{7}{l}{\textit{Other Ablations}} \\
          \midrule
          + Exploration Bonus      & 1395 & 48.6 & 48.9 & 1070 & 60.0 & 82.2 \\
          + Exploration \& Usefulness    & 1257 & \textbf{50.8}  & 51.1 & 1115 & 66.0 & \textbf{82.3} \\
          + KL Term              & 1474 & 48.1 & 55.0 & 1003 & 50.5 & 78.6 \\
          $-$ Weighting            & 1445 & 49.3 & 66.1 & 1165 & 72.1 & 81.4 \\
          Ordering 1 & 1638 & 49.1 & 51.4 & 1181 & 73.9 & 80.9 \\
          Ordering 2 & 1548 & 49.7 & 62.1 & 1237 & 79.6 & 80.2 \\
          Ordering 3 & 1421 & 50.8 & 72.1 & 1253 & 81.0 & 80.9 \\
          \midrule
          \rowcolor{red!7} \multicolumn{7}{l}{\textit{Full Method}} \\
          \midrule
          \textbf{RcRL (Ours)}     & \textbf{1577} & 50.1 & \textbf{69.3} & \textbf{1181} & \textbf{73.9} & 80.9 \\
          \bottomrule
        \end{tabular}%
        }
      \end{sc}
    \end{small}
  \end{center}
  \label{tab:ablations}
\end{table*}

\subsection{Curriculum Pacing}
\label{subsec:curriculum_pacing}
\begin{figure}[h]
  \centering
  \includegraphics[width=0.9\columnwidth]{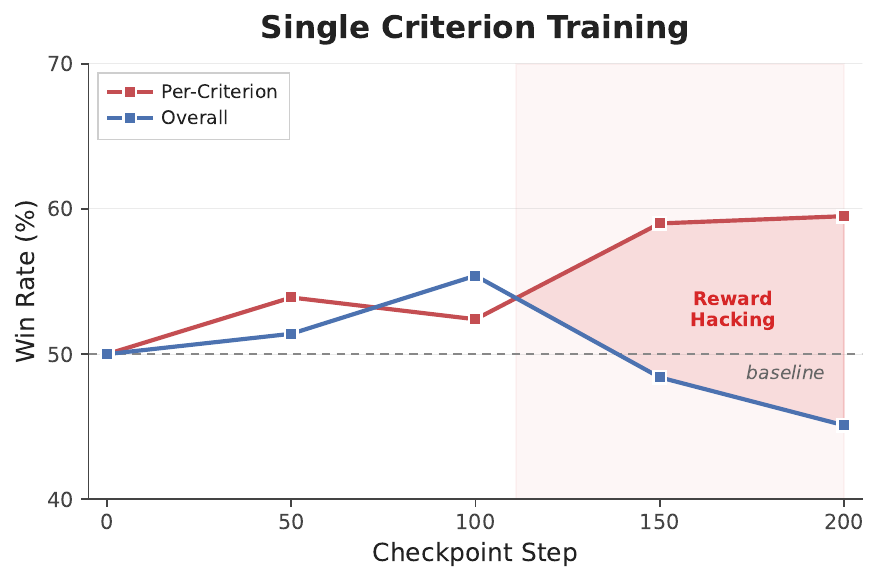}
  \caption{\textbf{Single-criterion training dynamics.} Win rates averaged across five models, each trained on a single criterion (e.g., plot coherence, character development). Per-criterion performance (red) improves steadily, while overall story quality (blue) peaks around step 100 before declining.}
  \label{fig:reward-hacking-per-criterion}
  \vspace{-1em}
\end{figure}

A key design choice in RcRL is when to introduce a new criterion to the curriculum. In Figure \ref{fig:reward-hacking-per-criterion} we train on a single criterion for 200 steps, then evaluate periodically on overall story quality and the specific criterion the model was trained on. Averaged across four criteria, the overall win rate climbs until approximately step 100 before declining, motivating $n=100$ for RcRL. Advancing the curriculum at this inflection point allows RcRL to capture each criterion's benefits during the informative phase, then shift the optimization target before degradation occurs.

\subsection{Entropy Preservation}
\begin{figure}[t]
  \centering
  \includegraphics[width=0.933\columnwidth]{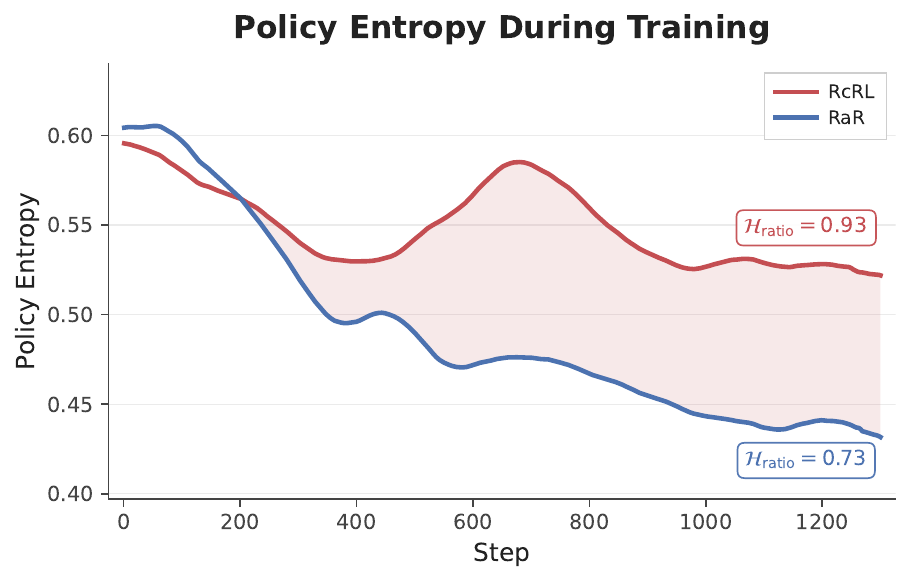}
  \caption{\textbf{Policy entropy during training.} The static rubric baseline exhibits a steady decline in entropy, indicating mode collapse toward a narrow set of outputs. RcRL maintains consistently higher entropy throughout training.}
  \label{fig:entropy}
  \vspace{-1em}
\end{figure}

Figure \ref{fig:entropy} shows that RaR exhibits steady entropy decline, which manifests in concrete output pathologies: RaR-trained models exhibit elevated n-gram repetition and, in select stories, the insertion of non-English tokens (Appendix~\ref{app:failure-analysis}). In contrast, RcRL maintains near-constant entropy throughout training with $\mathcal{H}_{\text{ratio}} = 0.93$ (ratio of final to initial entropy). This stability arises from (i) the curriculum continuously reshaping the reward landscape by updating the objective before the model can overfit to previous ones, and (ii) pairwise rewards that force the judge to discriminate regardless of output quality, preventing the vanishing gradients that accompany absolute reward saturation. Together, these mechanisms sustain exploration capacity, allowing the model to find diverse writing strategies.

\subsection{Cross-Domain Applicability}
\label{sec:cross-domain}

While our primary experiments focus on creative writing, RcRL's core mechanism is not specific to any domain. To test the generalizability of RcRL, we apply the method to HealthBench \citep{arora2025healthbench}, a health conversation/advising benchmark with per-example evaluation rubrics. We train Qwen3-4B with a fixed set of 13 general-purpose criteria constructed from HealthBench's evaluation rubrics (provided in Appendix \ref{app:healthbench-rubric}), using the same hyperparameters as in creative writing with the exception of $n=50$ steps per curriculum phase. We note that HealthBench provides instance-specific rubrics but we do not need these, and RcRL with a fixed rubric is competitive even against instance-specific rubrics.

We showcase two sets of results: one trained and evaluated on the full HealthBench split dataset and another trained and evaluated on the split consensus subset, where each per-example rubric criterion has been independently validated by multiple medical professionals so the criteria themselves are higher-quality and less noisy. On the full HealthBench dataset, RcRL achieves a 14.60\% score improvement over the base model compared to 8.26\% for RaR; on the consensus subset, RcRL achieves a 25.38\% improvement versus 7.53\% for RaR (Table~\ref{tab:healthbench}). These gains confirm the results observed in our creative writing evaluations.

\begin{table}[h]
  \caption{\textbf{HealthBench cross-domain results.} Score improvement over Qwen 3 4B on full HealthBench and the consensus subset (criteria validated by multiple annotators). Best in \textbf{bold}.}
  \begin{center}
    \begin{small}
      \begin{sc}
        \resizebox{\columnwidth}{!}{%
        \begin{tabular}{l cc}
          \toprule
          Method & HealthBench $\Delta$ & Consensus $\Delta$ \\
          \midrule
          Qwen 3 4B (Baseline) & --   & --   \\
          Qwen 3 4B RaR    & +8.26\% & +7.53\% \\
          Instance-specific & +13.52\% & +10.52\% \\
          Absolute RcRL    & +13.40\% & +23.87\% \\
          \rowcolor{red!7} \textbf{RcRL (Ours)} & \textbf{+14.60\%} & \textbf{+25.38\%} \\
          \bottomrule
        \end{tabular}%
        }
      \end{sc}
    \end{small}
  \end{center}
  \label{tab:healthbench}
\end{table}

Notably, the relative contribution of RcRL's two mechanisms differs across domains. In creative writing, the Pairwise RaR baseline (no curriculum) reaches 69.6\% on FictionBench (Table~\ref{tab:ablations}), only modestly below RcRL, indicating that pairwise rewards provide significant improvements. On HealthBench, however, the Absolute RcRL configuration (curriculum without pairwise rewards) closes most of the gap to full RcRL (+13.40\% vs +14.60\% on the full benchmark; +23.87\% vs +25.38\% on consensus).

To explain this, we observe that HealthBench criteria are more easily verifiable (e.g.\ ``does the response list all side effects of any suggested treatment?''), whereas creative-writing criteria are subjective (e.g.\ ``does the story have a clear plot structure?''). Verifiable criteria naturally admit calibrated absolute scores, so pairwise rewards extract less additional ranking signal beyond what absolute scores already provide. The complementarity of the curriculum and pairwise rewards is what makes RcRL robust across domains with structurally different reward signals.

\definecolor{colorours}{HTML}{c92c2c}
\definecolor{colorbaseline}{HTML}{2878b5}

\newcommand{\prefbar}[2]{%
  % #1 = baseline percentage (left), #2 = ours percentage (right)
  \begin{tikzpicture}[baseline=-0.3em]
    \def\barwidth{3.8cm}
    \def\barheight{0.38cm}
    \def\baseline{#1}
    \def\ours{#2}
    
    % Calculate widths
    \pgfmathsetmacro{\basewidth}{\baseline/100*\barwidth}
    \pgfmathsetmacro{\ourswidth}{\ours/100*\barwidth}
    
    % Baseline bar (left, blue) - transparent fill with solid outline
    \fill[colorbaseline, opacity=0.15] (0,0) rectangle (\basewidth pt,\barheight);
    \draw[colorbaseline, line width=1.2pt] (0,0) rectangle (\basewidth pt,\barheight);
    
    % Ours bar (right, red) - transparent fill with solid outline
    \fill[colorours, opacity=0.15] (\basewidth pt,0) rectangle (\barwidth,\barheight);
    \draw[colorours, line width=1.2pt] (\basewidth pt,0) rectangle (\barwidth,\barheight);
    
    % 50% marker line
    \draw[black!50, line width=1pt, dashed] (0.5*\barwidth,0) -- (0.5*\barwidth,\barheight);
    
    % Confidence interval shading (gray box around the win rate)
    % \pgfmathsetmacro{\cilow}{max(\basewidth,(\ours-5)/100*\barwidth)}
    % \pgfmathsetmacro{\cihigh}{min(\barwidth,(\ours+5)/100*\barwidth)}
    % \fill[black, opacity=0.2] (\cilow pt,0) rectangle (\cihigh pt,\barheight);
    
    % Win rate percentage label
    \node[anchor=south west, font=\scriptsize\bfseries, text=colorours] 
      at (\barwidth+0.08cm, 0.02cm) {#2\%};
  \end{tikzpicture}%
}

\section{Discussion}

\subsection{Limitations}
Evaluation in non-verifiable domains remains fundamentally noisy. On creative writing, even among domain-expert human annotators, inter-rater agreement reaches only moderate levels ($\kappa \approx 0.6$) \citep{gomez-rodriguez-williams-2023-confederacy}, and rubric-based judgments in more verifiable domains like HealthBench still depend on judge model fidelity. We mitigate stochastic noise and judge bias via hundreds of comparisons across multiple frontier judges and observe consistent performance across all settings.

The curriculum with pairwise rewards scales quadratically with rollout batch size and linearly with training steps. For our experiments with 8 rollouts and a 13-criterion curriculum this remains tractable, but scaling to larger $G$ or finer-grained rubrics would require approximations such as tournament-style evaluation. Under the Bradley-Terry model, $\mathcal{O}(G \log G)$ adaptive comparisons (e.g., Swiss-system tournaments) can recover the ordinal ranking while preserving expected win rates, suggesting much of the gain can be retained at a substantially reduced cost.

\subsection{Future Work}
\label{sec:futurework}
RcRL advances the curriculum every 100 steps uniformly; learning to detect per-criterion inflection points automatically could improve both efficiency and final performance. RcRL currently requires a hand-specified rubric; automating criterion discovery from preference data would reduce the cost of applying the method to new domains. Scaling to substantially larger rubrics will additionally require adaptive or stochastic curriculum strategies, since fixed-order presentation becomes less defensible with many criteria. Additionally, verification compute can be scaled significantly, which might improve the robustness further \citep{singh2026v1unifyinggenerationselfverification, venkatraman2026recursiveselfaggregationunlocksdeep}. Many long-horizon tasks, such as research or multi-step planning, produce outputs that cannot be verified, and rubric rewards are already used to train such agents \citep{goel2025trainingaicoscientistsusing}. Extending RcRL to these settings is a natural next step, and we describe our planned approach below.

\paragraph{Extension to long-horizon tasks.}
In long-horizon open-ended tasks such as deep research \citep{shao2026drtulureinforcementlearning} and research planning \citep{goel2025trainingaicoscientistsusing}, the problems motivating RcRL become more important since a single rubric score must summarize many intermediate decisions as well as many criteria, and arrives only at the end of a long trajectory. RcRL can be extended to these settings by aligning the rubric curriculum with the task horizon. Suppose a task consists of $H$ stages, each producing an intermediate output that the next stage builds on (e.g., a set of sources, a synthesis of them, and a final report), and let $\mathcal{R}_h$ contain the criteria that assess the output of stage $h$. We can order the rubric by stage, $\mathcal{R} = (\mathcal{R}_1,\ldots,\mathcal{R}_H)$, and rather than performing all $H$ stages and judging only the final output throughout training, the policy performs only the first $h$ stages while criteria from $\mathcal{R}_h$ are being added, and the judge compares the outputs of stage $h$. The rubric and the task horizon can thus grow together, combining the curriculum over criteria with one over horizon length \citep{h1paper}, while the reward is computed exactly as in RcRL so that earlier criteria remain in the reward.

\subsection{Conclusion}
We introduce Rubric Curriculum RL (RcRL), a method that combines pairwise rewards with curriculum learning to provide training signal for improving performance in non-verifiable domains. On creative writing, RcRL achieves a 73.9\% win rate on FictionBench and a 70.5\% human preference over its base model, without human preference data or stronger teacher models. On HealthBench, RcRL improves over the base model by 14.6\%, outperforming all training baselines including instance-specific rubrics. Improvements scale with model size, suggesting the approach becomes increasingly effective at frontier scales. Our results are amongst the first to demonstrate that in non-verifiable domains where verification is easier than generation, it is possible to use this asymmetry as training signal for scalable self-improvement.

\section*{Impact Statement}
This paper presents work whose goal is to advance the field of Machine
Learning. It serves to improve RL post-training performance on an important, understudied domain.

\bibliography{main}
\bibliographystyle{icml2026}

\newpage
\appendix
\onecolumn
\section{Appendix}

\subsection{Decomposing and exploiting the generation--verification gap}
\label{app:theory}

In this section, our goal is to formalize the generation-verification (GV) gap and show that Claim 1 (Gap exists) in Section \ref{sec:gv_gap} implies Claim 3 (Gap is exploitable). We first decompose the GV gap into two terms that describe variation in response quality and the judge's ability to rank responses. We then show that a positive best-of-two gap implies a local improvement in generation quality under an idealized KL-regularized policy update.

\paragraph{The pairwise GV gap.}
Let $Q(x,y)\in[0,1]$ denote response quality, and define
\[
  \Delta(y,y'\mid x):=Q(x,y)-Q(x,y').
\]
Let $\hat{p}(y\succ y'\mid x)$ be the probability that the judge ranks $y$ above $y'$ on prompt $x$. We assume that comparison probabilities satisfy
\[
  \hat{p}(y\succ y'\mid x)+\hat{p}(y'\succ y\mid x)=1,
\]
with judge ties resolved uniformly at random. As in RcRL, the judge is the frozen initial model $\pi_{\mathrm{judge}}=\pi_0$, while the generation policy $\pi$ may change during training.

Define the pairwise GV gap by
\begin{equation}
  \gap(\pi):=\mathbb{E}_{x\sim\mathcal{X},\,y,y'\sim\pi(\cdot\mid x)}
  \left[\Delta(y,y'\mid x)\hat{p}(y\succ y'\mid x)\right],
  \label{eq:app-pairwise-gap}
\end{equation}
where $y$ and $y'$ are independent conditional on $x$. Unless otherwise stated, expectations below use this distribution over prompts and response pairs.

This is the expected quality gain from using the judge to select between two independent responses. Writing $\bar{V}^{(2)}_\pi$ for the expected quality of the selected response and $\bar{G}(\pi)$ for the expected quality of a single response, we have
\begin{align*}
  \bar{V}^{(2)}_\pi-\bar{G}(\pi)
  &= \mathbb{E}\Big[Q(x,y)\hat{p}(y\succ y'\mid x)+Q(x,y')\bigl(1-\hat{p}(y\succ y'\mid x)\bigr)\Big]-\mathbb{E}[Q(x,y)]\\
  &= \mathbb{E}\Big[\Delta(y,y'\mid x)\hat{p}(y\succ y'\mid x)\Big]+\mathbb{E}[Q(x,y')]-\mathbb{E}[Q(x,y)]\\
  &= \gap(\pi).
\end{align*}
The final equality follows because $y$ and $y'$ have the same conditional distribution. When the judge is the generation model itself, this agrees with the $N=2$ case of the GV gap in the main text. Allowing a separate frozen judge extends the definition to later stages of training.

\paragraph{Quality diversity and judge skill.}
For a pair $(y,y')$, define the quality-ordered pair $(y^+,y^-)=(y,y')$ if $Q(x,y)>Q(x,y')$, and $(y^+,y^-)=(y',y)$ otherwise. Write
\[
  q(y,y'):=\hat{p}(y^+\succ y^-\mid x),
\]
suppressing the dependence on $x$. For responses of unequal quality, $q$ is the probability that the judge ranks the higher-quality response above the lower-quality one. The ordering of equal-quality responses does not affect the analysis because $\Delta=0$ for such pairs.

\begin{proposition}
The GV gap factorizes as
\[
  \gap(\pi)=\mu(\pi)\nu(\pi),
\]
where
\[
  \mu(\pi):=\mathbb{E}[|\Delta|]
\]
is the policy's \emph{quality diversity}, and, when $\mu(\pi)>0$,
\[
  \nu(\pi):=\frac{\mathbb{E}\bigl[|\Delta|(q-\tfrac12)\bigr]}{\mathbb{E}[|\Delta|]}
\]
is the judge's $|\Delta|$-weighted \emph{ranking skill}, centered around chance. When $\mu(\pi)=0$, set $\nu(\pi)=0$.
\end{proposition}

\begin{proof}
Since $\mathbb{E}[\Delta]=0$,
\[
  \gap(\pi)=\mathbb{E}\left[\Delta\left(\hat{p}(y\succ y'\mid x)-\tfrac12\right)\right].
\]
We show that
\[
  \gap(\pi)=\mathbb{E}\bigl[|\Delta|(q-\tfrac12)\bigr].
\]

If $Q(x,y)>Q(x,y')$, then $\Delta=|\Delta|$ and $q=\hat{p}(y\succ y'\mid x)$, so
\[
  \Delta\left(\hat{p}(y\succ y'\mid x)-\tfrac12\right)=|\Delta|(q-\tfrac12).
\]
If $Q(x,y)<Q(x,y')$, then $\Delta=-|\Delta|$ and
\[
  q=\hat{p}(y'\succ y\mid x)=1-\hat{p}(y\succ y'\mid x).
\]
Hence,
\[
  \Delta\left(\hat{p}(y\succ y'\mid x)-\tfrac12\right)=-|\Delta|(1-q-\tfrac12)=|\Delta|(q-\tfrac12).
\]
When the two responses have equal quality, both sides are zero. Taking expectations gives our result.

For $\mu(\pi)>0$, multiplying and dividing by $\mathbb{E}[|\Delta|]$ yields $\gap(\pi)=\mu(\pi)\nu(\pi)$. If $\mu(\pi)=0$, then $\Delta=0$ almost surely, so $\gap(\pi)=0$ and the identity also holds.
\end{proof}

The term ``$|\Delta|$-weighting'' of the judge skill $(q-\tfrac12)$ is justified because, when $\mu(\pi)>0$, the weights $|\Delta|/\mathbb{E}[|\Delta|]$ are nonnegative and have expectation one, while the term ``quality diversity'' comes from noticing that a deterministic policy has $\mu(\pi)=0$, while a policy generating outputs of different quality with positive probability has $\mu(\pi)>0$. In particular, the GV gap is positive if and only if the judge has positive quality-weighted ranking skill and the policy has positive quality diversity. The decomposition distinguishes ways in which the GV gap can vanish or become negative over the course of training:\begin{enumerate}
  \item[(a)]
  \emph{Diversity collapse ($\mu\to0$).}
  The policy's responses cease to differ in quality, whether through mode collapse or convergence to responses of similar quality.

  \item[(b)]
  \emph{Judge skill limit ($\nu\to0$).}
Differences persist ($\mu>0$), but the judge's quality-weighted ranking advantage vanishes. This can occur when $q\to\tfrac12$ on quality-differing output pairs, meaning the judge is no longer able to distinguish between them, or when correct and incorrect preferences cancel in the weighted average.
  \item[(c)]
  \emph{Goodharting ($\nu<0$).}
  The judge favors lower-quality responses on average after weighting by quality differences. This can occur when optimization shifts the policy toward responses that exploit errors in the judge.
\end{enumerate}

These cases motivate maintaining quality diversity, improving the judge's discriminative signal, and changing evaluation criteria to limit overoptimization of a fixed proxy.

\paragraph{From a positive GV gap to self-improvement.}
The GV gap is exactly the true-quality gain from using the best-of-$2$ sampling policy $\pi^{\mathrm{bo2}}$ (sample two responses from $\pi$ and return the judge's preferred one) instead of $\pi$, namely $\bar{G}(\pi^{\mathrm{bo2}})-\bar{G}(\pi)=\gap(\pi)$. So a positive GV gap already implies self-improvement via sampling. We now show that it also implies local self-improvement via RL. We formalise this by assuming that, for each prompt $x$, the policy is updated via the idealised KL-regularised RL update
\begin{equation}\label{eq:rl-update}
  \pi_{t+1}:=\operatorname*{arg\,max}_{\pi}\left\{\mathbb{E}_{\pi}[R]-\tfrac{1}{\beta}\kl{\pi}{\pi_t}\right\}\propto\pi_t e^{\beta R}\,,\qquad\beta>0\,,
\end{equation}
which GRPO-style methods approximate. We take $R=R_{\pi_t}$ to be the pairwise win-rate reward against the current policy,
$R_{\pi_t}(y):=\mathbb{E}_{y'\sim\pi_t}[\hat{p}(y\succ y')]$,
held fixed during the update. In RcRL, a weighted version is estimated from a group of $G$ rollouts as
\[
  R_{\mathrm{W\text{-}pair}}(y_i)=\sum_m w^{(m)}\,\frac{1}{G-1}\sum_{j\ne i}y^{(m)}_{i\succ j}\,.
\]
We obtain the following result, which shows self-improvement provided $\gap(\pi_t)>0$ and $\beta$ is small.

\begin{proposition}
If $R=R_{\pi_t}$ is the pairwise win-rate reward, then the idealised KL-regularised update yields
\[
  \bar{G}(\pi_{t+1})-\bar{G}(\pi_t)=\frac{\beta}{2}\gap(\pi_t)+O(\beta^2)\quad\text{as }\beta\to0^+\,.
\]
If the reward is weighted as $R=\sum_m w^{(m)}R^{(m)}_{\pi_t}$, the leading term becomes $\frac{\beta}{2}\sum_m w^{(m)}\gap_m(\pi_t)$, where $\gap_m$ is defined via the rubric-specific judge $\hat{p}_m$ (also assumed complementary).
\end{proposition}

\begin{proof}
We can fix $x$ and omit it; the prompt-averaged result follows by averaging over $x\sim\mathcal{X}$, since all quantities below are uniformly bounded. We first relate the quality improvement to the covariance between quality and reward. For any bounded reward $R$, let $\pi_\beta=\pi e^{\beta R}/Z_\beta$ with $Z_\beta=\mathbb{E}_\pi[e^{\beta R}]$. Since $d\pi_\beta/d\pi=e^{\beta R}/Z_\beta$, for any integrable $f$ we have $\mathbb{E}_{\pi_\beta}[f]=\mathbb{E}_\pi[fe^{\beta R}]/Z_\beta$. Writing $A_\beta=\mathbb{E}_\pi[Qe^{\beta R}]$, the quotient rule gives
\[
  \frac{d}{d\beta}\mathbb{E}_{\pi_\beta}[Q]=\frac{A'_\beta}{Z_\beta}-\frac{A_\beta}{Z_\beta}\frac{Z'_\beta}{Z_\beta}=\frac{\mathbb{E}_\pi[QRe^{\beta R}]}{Z_\beta}-\frac{\mathbb{E}_\pi[Qe^{\beta R}]}{Z_\beta}\frac{\mathbb{E}_\pi[Re^{\beta R}]}{Z_\beta}=\Cov_{\pi_\beta}(Q,R)\,.
\]
Since $Q$ and $R$ are bounded, so is $\frac{d}{d\beta}\Cov_{\pi_\beta}(Q,R)$, and Taylor expansion around $\beta=0$ with $\pi=\pi_t$, $\pi_\beta=\pi_{t+1}$ gives
\[
  \bar{G}(\pi_{t+1})-\bar{G}(\pi_t)=\beta\Cov_{\pi_t}(Q,R)+O(\beta^2)\,.
\]
It remains only to relate the covariance to the GV gap. Exchanging $y\leftrightarrow y'$ and using complementarity gives $\mathbb{E}_y[R_\pi(y)]=\mathbb{E}_{y,y'}[\hat{p}(y\succ y')]=\mathbb{E}_{y,y'}[\hat{p}(y'\succ y)]=1-\mathbb{E}_{y,y'}[\hat{p}(y\succ y')]$, so $\mathbb{E}_y[R_\pi(y)]=\frac12$. It follows, interchanging $y\leftrightarrow y'$ in the third line and using complementarity in the fourth, that
\begin{align*}
  \Cov_\pi(Q,R_\pi)&=\mathbb{E}_y[Q(y)R_\pi(y)]-\mathbb{E}_y[Q(y)]\,\mathbb{E}_y[R_\pi(y)]\\
  &=\mathbb{E}_{y,y'}[Q(y)\hat{p}(y\succ y')]-\tfrac12\mathbb{E}_y[Q(y)]\\
  &=\tfrac12\mathbb{E}_{y,y'}[Q(y)\hat{p}(y\succ y')+Q(y')\hat{p}(y'\succ y)]-\tfrac12\mathbb{E}_y[Q(y)]\\
  &=\tfrac12\mathbb{E}_{y,y'}[(Q(y)-Q(y'))\hat{p}(y\succ y')+Q(y')]-\tfrac12\mathbb{E}_y[Q(y)]=\tfrac12\gap(\pi)\,.
\end{align*}
\end{proof}

This is a local version of Claim~1 $\Rightarrow$ Claim~3, and it requires a positive gap at the current policy $\pi_t$ (for the weighted reward, $\sum_m w^{(m)}\gap_m(\pi_t)>0$). Rather than a guarantee for practical GRPO training, it formalizes the mechanism behind our approach and identifies, via $\gap=\mu\nu$, what conditions are useful for self-improvement.

\subsection{Why Criterion-Wise Rewards Improve Credit Assignment}
\label{app:pairwise-theory}

Now, we informally analyze the effect of judging one rubric criterion at a time over one holistic reward over the full rubric. Since the win-rate reward is built from individual binary comparisons, we analyze a single comparison and ask how much it reveals about the quality gap on the target criterion.
\paragraph{Setup.}
Consider two rollouts $\mathbf{o}_i,\mathbf{o}_j$ and a target rubric criterion $m$.
Let
\[
\Delta^{(m)}_{ij}
:=
\nu^{(m)}(\mathbf{o}_i)-\nu^{(m)}(\mathbf{o}_j)
\]
denote the latent quality gap between the two rollouts on criterion $m$, where
$\nu^{(m)}(\mathbf{o})$ is the latent quality of rollout $\mathbf{o}$ on that criterion.
We use $\Delta_m := \Delta^{(m)}_{ij}$ for the sake of notational simplicity.

A criterion-wise judge compares the two rollouts only on criterion $m$. We model the
probability that it prefers $\mathbf{o}_i$ over $\mathbf{o}_j$ using the standard
Bradley-Terry form
\[
p_{\mathrm{crit}}
=
\Pr(\mathbf{o}_i \succ \mathbf{o}_j \mid r^{(m)})
=
\sigma(\beta \Delta_m),
\]

where $\sigma(z)=(1+e^{-z})^{-1}$ and $\beta>0$ controls the slope of the preference model: larger $\beta$ makes the preference probability more sensitive to a fixed nonzero quality gap, while smaller $\beta$ leaves the comparison probability closer to $\tfrac12$.

\paragraph{Holistic versus criterion-wise comparisons.}
A holistic judge instead compresses all $K$ rubric criteria into one verdict. Suppose
the holistic verdict implicitly weights criterion $\ell$ by $\alpha_\ell$, with
$\alpha_\ell\ge 0$ and $\sum_{\ell=1}^K \alpha_\ell=1$. The holistic quality gap can be
written as
\[
\sum_{\ell=1}^K \alpha_\ell \Delta^{(\ell)}_{ij}
=
\alpha_m \Delta_m + Z_m,
\]
where
\[
Z_m
:=
\sum_{\ell\ne m} \alpha_\ell \Delta^{(\ell)}_{ij}.
\]
Here, $Z_m$ is the total contribution of all non-target criteria to the holistic comparison. That is, the contribution of everything other than $m$ to the holistic comparison. If $m$ is originality, for instance, $Z_m$ aggregates the weighted differences between the two rollouts in coherence, style, instruction following, and the remaining rubric items. The holistic comparison is driven by both the  target gap $\Delta_m$ and the aggregate non-target gap $Z_m$, so depending on its sign and magnitude, $Z_m$ can reinforce, cancel, or reverse the preference that criterion $m$ would produce on its own. Under the same $\beta$, the holistic comparison probability is
\[
p_{\mathrm{hol}}
=
\sigma\!\left(\beta(\alpha_m\Delta_m+Z_m)\right).
\]
Thus, in a holistic comparison, the target criterion affects the judge's decision only
through the attenuated term $\alpha_m\Delta_m$.

\begin{theorem}[Criterion dilution]
For a single Bernoulli comparison, the Fisher information about the target gap
$\Delta_m$ is
\[
I=\frac{(p')^2}{p(1-p)}.
\]
Therefore,
\[
I_{\mathrm{hol}}
=
\beta^2\alpha_m^2\,p_{\mathrm{hol}}(1-p_{\mathrm{hol}}),
\qquad
I_{\mathrm{crit}}
=
\beta^2\,p_{\mathrm{crit}}(1-p_{\mathrm{crit}}),
\]
and the exact information ratio is
\[
\frac{I_{\mathrm{hol}}}{I_{\mathrm{crit}}}
=
\alpha_m^2
\frac{p_{\mathrm{hol}}(1-p_{\mathrm{hol}})}
     {p_{\mathrm{crit}}(1-p_{\mathrm{crit}})}.
\]
\end{theorem}

\begin{proof}
For a Bernoulli outcome with success probability $p(\Delta_m)$, the Fisher information
about $\Delta_m$ is $(p')^2/[p(1-p)]$. For the holistic comparison,
\[
p'_{\mathrm{hol}}
=
\beta\alpha_m\,p_{\mathrm{hol}}(1-p_{\mathrm{hol}}),
\]
while for the criterion-wise comparison,
\[
p'_{\mathrm{crit}}
=
\beta\,p_{\mathrm{crit}}(1-p_{\mathrm{crit}}).
\]
Substituting these derivatives into the Bernoulli Fisher information formula gives the expressions above and their ratio.
\end{proof}

The exact ratio combines two effects: the structural factor $\alpha_m^2$, through which
the target criterion enters a holistic comparison, and a modulation by the Bernoulli
variances $p_{\mathrm{hol}}(1-p_{\mathrm{hol}})$ and
$p_{\mathrm{crit}}(1-p_{\mathrm{crit}})$. Whenever neither comparison is saturated, this
modulation is bounded: if $p_{\mathrm{hol}},p_{\mathrm{crit}}\in[\delta,1-\delta]$ for
some $\delta\in(0,\tfrac12]$, then $p(1-p)\in[\delta(1-\delta),\tfrac14]$ for both, and
\[
4\delta(1-\delta)\,\alpha_m^2
\;\le\;
\frac{I_{\mathrm{hol}}}{I_{\mathrm{crit}}}
\;\le\;
\frac{\alpha_m^2}{4\delta(1-\delta)},
\]
so the ratio stays within a constant factor of $\alpha_m^2$. The tightest case is
$\delta=\tfrac12$ which is the non-saturated case, where both comparisons are at the decision boundary
($p_{\mathrm{hol}}=p_{\mathrm{crit}}=\tfrac12$), the factor equals one, and the ratio is
exactly $\alpha_m^2$. For an equally weighted rubric ($\alpha_m=1/K$), the dilution is of
order $1/K^2$ per criterion.

This bound is optimistic for the holistic judge, since it treats $Z_m$ as fixed. The
learner observes only the single holistic preference, not how it is split into
$\alpha_m\Delta_m$ and $Z_m$, so a holistic label cannot on its own attribute the
preference to the target criterion rather than the others. The criterion-wise rewards used in our method remove
this ambiguity by comparing rollouts directly on the criterion being optimized.

\paragraph{Relative versus absolute rewards.}
The reward in RcRL is the within-group win rate built from these pairwise outcomes, so it
depends on the judge only through the order it assigns to rollouts, not the magnitude of any
score. This means that it is invariant to any strictly increasing rescaling of the
judge's scores, and therefore requires only ordinal judgments rather than a magnitude-calibrated
scale; and its group mean equals $\tfrac12$ in expectation for every policy, since the logistic
model gives $\sigma(\beta\Delta)+\sigma(-\beta\Delta)=1$, so the mean reward cannot drift upward
and all signal comes from relative within-group differences. An absolute reward instead uses the cardinal score and exposes a fixed target that can be driven upward independently of quality,
the failure mode in Figure~\ref{fig:RaR-reward-hacking}.

\subsection{Cognitive Science Perspective on the Generation-Verification Gap}
\label{app:cogsci}

We sketch here an informal cognitive-science argument for why the verification-generation asymmetry is amplified in creative domains. Evaluation is generically easier than generation across most domains because generation involves search over a vast space of possibilities whereas evaluation does not. For creative tasks specifically, this asymmetry is amplified by the dual imperatives of creativity: outputs must be both \emph{good} (high quality) and \emph{novel} (different from existing work). Novelty inherently compounds the search problem, as the model must navigate away from familiar solutions while maintaining quality \citep{newell1962processes, anderson1972recognition}.

This perspective aligns with computational complexity intuitions. Verification requires compressing high-dimensional outputs into low-dimensional quality judgments, while generation requires constructing outputs that satisfy multiple simultaneous constraints. More formally, verification corresponds to checking whether a given candidate is a solution, whereas generation corresponds to finding one. Any verifier yields a generator by proposing candidates and testing each one, so the gap between the two lies in the cost of search, and whether this search can in general be avoided is the P versus NP question \citep{cook2023complexity}. In non-verifiable domains the verifier is an approximate quality judgment rather than an exact check, so selecting outputs with it raises expected quality rather than guaranteeing a correct output.

\subsection{Reward Hacking Analysis}
\label{app:repetition}
\label{app:failure-analysis}
Reward hacking in language model fine-tuning often manifests as increased n-gram repetition, where models exploit judge insensitivity to lexical redundancy \citep{holtzman2020curious}. Table~\ref{tab:repetition} reports the n-gram repetition score from \cite{paech2025slopScore}. RcRL exhibits consistently lower repetition rates than RaR, suggesting that curriculum learning and pairwise rewards preserve output diversity rather than collapsing to repetitive patterns that inflate scores.

\begin{table}[h]
\centering
\small
\caption{$N$-gram repetition rates (lower is better).}
\begin{tabular}{lccc}
\toprule
& $n$-gram repetition score \\
\midrule
Qwen3-4B & 11.51 \\
Qwen3-4B RaR & 13.58  \\
RcRL & 7.52 \\
\bottomrule
\end{tabular}

\label{tab:repetition}
\end{table}

Models that are trained directly with self-verification signal also exhibit reward hacking in the form of non-English tokens inserted into stories. An example of this is below.

\begin{tcolorbox}[
  colback=gray!5, colframe=gray!50,
  title=Sample Lines from RaR Story (with reward hacking artifact highlighted),
  breakable,
  fontupper=\small
]
\textbf{The Clouds Have a Language}

\medskip

Clara Morgan had covered small-town oddities for \textcolor{red}{\textbf{\begin{CJK}{UTF8}{gbsn}二十年\end{CJK}}}---lost heirlooms, missing cats, the man who claimed to speak to his deceased father. But nothing prepared her for Eli Cross.

\medskip

Eli was a relic of the 1950s, a retired schoolteacher with a beard like a tangled brush and eyes that seemed to hold the sky itself.
\end{tcolorbox}

\subsection{Additional Related Work}
We note that Compute as Teacher (CaT; \citealp{jayalath2026computeteacherturninginference}) implicitly use the generator-verifier gap in non-verifiable domains. In the setting where no ground truth answers are available, their work generates a pseudo-reference answer at every step of training. This step implicitly involves verifying the generated rollouts and identifying the points of disagreement between them before they can be resolved to produce a unified pseudo-reference answer. In non-verifiable domains, CaT generates a rubric that characterises the pseudo-reference answer and rewards rollouts by how many of the rubric criteria they satisfy according to a judge model, which can be treated as an implicit curriculum. Concurrently, EVOLM \citep{li2026evolmselfevolvinglanguagemodels} learns instance-specific discriminative rubrics that co-evolve with the policy, showing that rubric generation itself can be trained. RcRL assumes rubric criteria are available and focuses on the reward-design problem instead. In contrast, \citep{jia2025writingzerobridgegapnonverifiable} trains a separate pairwise reward model that requires a dataset of human preferences and a teacher model.

\section{Auxiliary Reward Bonuses}
\label{app:auxiliary-rewards}

We augment RcRL with two optional exploration bonuses: a \textbf{representation-based diversity bonus} (RepExp) adapted from \citet{tuyls2025representationbasedexplorationlanguagemodels}, and a \textbf{usefulness bonus} inspired by compression-based theories of learning \citep{schmidhuber2010formal}. Both are computed per-rollout, normalized via z-score standardization, and added to the pairwise reward with coefficient $\lambda = 0.1$.

\subsection{Representation-Based Exploration (RepExp)}

We encourage diverse rollouts by rewarding responses that occupy novel regions in the model's latent representation space.

\paragraph{Representation extraction \citep{tuyls2025representationbasedexplorationlanguagemodels}.} We follow the same RepExp method in this paper. For a response $y = (y^1, \ldots, y^T)$ to prompt $x$, we extract the mean-pooled final-layer hidden state over completion tokens only:
\begin{equation}
\tilde{h}_\theta(x, y) = \frac{1}{T} \sum_{t=1}^{T} h_\theta(x, y^{1:t}) \in \mathbb{R}^d,
\end{equation}
where $h_\theta(x, y^{1:t}) \in \mathbb{R}^d$ is the last-layer activation (prior to the unembedding matrix) after processing token $y^t$. We then L2-normalize to obtain unit-norm representations:
\begin{equation}
\bar{h}_\theta(x, y) = \frac{\tilde{h}_\theta(x, y)}{\|\tilde{h}_\theta(x, y)\|_2}.
\end{equation}

\paragraph{Dimensionality reduction.} To reduce computational cost, we apply a random projection from the model's hidden dimension $d$ to a lower dimension $d' = 256$. Let $P \in \mathbb{R}^{d \times d'}$ be a random Gaussian matrix with columns normalized to unit norm:
\begin{equation}
P_{:,j} \sim \mathcal{N}(0, I_d), \quad P_{:,j} \leftarrow \frac{P_{:,j}}{\|P_{:,j}\|_2}.
\end{equation}
The projected representation is $\bar{h}'_\theta(x, y) = \bar{h}_\theta(x, y)^\top P \in \mathbb{R}^{d'}$. By the Johnson-Lindenstrauss lemma, this approximately preserves pairwise distances with high probability.

\paragraph{Elliptical bonus.} Given a batch of $N$ rollouts $\{y_1, \ldots, y_N\}$ for prompt $x$, we compute the regularized empirical covariance in the projected space:
\begin{equation}
\Sigma = \lambda I_{d'} + \sum_{i=1}^{N} \bar{h}'_i (\bar{h}'_i)^\top \in \mathbb{R}^{d' \times d'},
\end{equation}
where $\bar{h}'_i := \bar{h}'_\theta(x, y_i)$ and $\lambda > 0$ is a regularization parameter ensuring invertibility. The RepExp bonus for rollout $y_i$ is its leverage score:
\begin{equation}
r_{\text{RepExp}}(y_i) = (\bar{h}'_i)^\top \Sigma^{-1} \bar{h}'_i.
\end{equation}

\paragraph{Interpretation.} The leverage score measures how much rollout $y_i$ influences the fit of a linear model trained on all rollout representations. Equivalently, it quantifies the novelty of $y_i$'s representation relative to the span of other rollouts. If $\bar{h}'_i$ lies in a direction well-covered by other representations, the corresponding eigenvalue of $\Sigma$ is large and the bonus is small. Conversely, if $\bar{h}'_i$ occupies an underrepresented direction, the bonus is large, encouraging the policy to explore diverse regions of output space.

\subsection{Usefulness Bonus}

Based on early work from \cite{schmidhuber2010formal}, we introduce a usefulness bonus based on the principle that informative rollouts are those that, when provided as in-context demonstrations, most reduce the model's uncertainty about other rollouts.

\paragraph{Log-perplexity computation.} For a completion $y$ given prompt $x$, define the mean per-token cross-entropy loss:
\begin{equation}
\mathcal{L}_\theta(y \mid x) = \frac{1}{|y|} \sum_{t=1}^{|y|} -\log p_\theta(y^t \mid x, y^{<t}),
\end{equation}
where $|y|$ denotes the number of tokens in $y$. This equals $\log \text{PPL}_\theta(y \mid x)$, the log-perplexity.

\paragraph{In-context perplexity reduction.} Given rollouts $\{y_1, \ldots, y_N\}$ for prompt $x$, we measure how much each rollout $y_i$ reduces the log-perplexity of other rollouts when provided as an in-context example. Let $x \oplus y_i$ denote the augmented prompt:
\begin{equation}
x \oplus y_i = x + \texttt{"\textbackslash nHere is an example story for you: "} + y_i.
\end{equation}
The usefulness of $y_i$ is the total reduction in log-perplexity across all other rollouts:
\begin{equation}
r_{\text{useful}}(y_i) = \sum_{j \neq i} \left[ \mathcal{L}_\theta(y_j \mid x) - \mathcal{L}_\theta(y_j \mid x \oplus y_i) \right].
\end{equation}

\paragraph{Interpretation.} A rollout that substantially reduces sibling perplexities captures generalizable structure such as clear narrative patterns, consistent style, or effective prompt interpretation that helps the model predict diverse completions. Idiosyncratic or degenerate outputs provide little predictive value and receive low usefulness scores. This connects to compression-based theories of learning: useful examples are those enabling efficient prediction of related data.

\subsection{Reward Normalization and Combination}

Both auxiliary rewards exhibit different scales across prompts. We normalize each via z-score normalization within each batch:
\begin{equation}
\hat{r}(y_i) = \frac{r(y_i) - \mu}{\sigma + \epsilon}, \quad \mu = \frac{1}{N}\sum_{j=1}^{N} r(y_j), \quad \sigma = \sqrt{\frac{1}{N}\sum_{j=1}^{N} (r(y_j) - \mu)^2},
\end{equation}
where $\epsilon = 10^{-6}$ prevents division by zero when all rewards are identical.

The final per-rollout reward combines the pairwise quality signal with normalized auxiliary bonuses:
\begin{equation}
r_i = r_{\text{pairwise}}(y_i) + \lambda_{\text{rep}} \cdot \hat{r}_{\text{RepExp}}(y_i) + \lambda_{\text{use}} \cdot \hat{r}_{\text{useful}}(y_i),
\end{equation}
where $\lambda_{\text{rep}}, \lambda_{\text{use}} \in \{0, 0.1\}$. We evaluate three configurations: baseline ($\lambda_{\text{rep}} = \lambda_{\text{use}} = 0$), RepExp only ($\lambda_{\text{rep}} = 0.1, \lambda_{\text{use}} = 0$), and both ($\lambda_{\text{rep}} = \lambda_{\text{use}} = 0.1$). The z-score normalization ensures both bonuses have zero mean and unit variance within each batch, allowing them to contribute comparably regardless of their raw magnitudes while preserving relative rankings.

\section{Experimental Setup and Curriculum Details}
\label{app:experimental-setup}

\subsection{Training Data}

We generate training prompts by having the base model analyze short stories and produce prompts with the goal of recreating the given story. This approach yields diverse, high-quality creative writing prompts that capture the complexity of real fiction. We use 200 synthetically generated unique prompts for training.

\subsection{Training Configuration}
\label{app:trainingconfig}
We train all models using the TRL library \citep{vonwerra2020trl} with Group Relative Policy Optimization (GRPO). Table~\ref{tab:training-config} summarizes the key hyperparameters for our primary Qwen3-4B experiments.

\begin{table}[h]
\centering
\small
\caption{\textbf{Training hyperparameters for RcRL.} Configuration for Qwen3-4B experiments. We use LoRA for parameter-efficient fine-tuning.}
\begin{tabular}{ll}
\toprule
\textbf{Parameter} & \textbf{Value} \\
\midrule
\multicolumn{2}{l}{\textit{GRPO Configuration}} \\
Learning rate & $5 \times 10^{-6}$ \\
LR scheduler & Cosine \\
Warmup steps & 50 \\
Per-device batch size & 2 \\
Gradient accumulation steps & 4 \\
Effective batch size & 8 \\
Number of rollouts ($G$) & 8 \\
Max prompt length & 768 tokens \\
Max completion length & 1024 tokens \\
Total training steps & 1300 \\
Loss type & Dr.\ GRPO \\
Precision & FP16 \\
\midrule
\multicolumn{2}{l}{\textit{Generation Settings}} \\
Top-$p$ & 0.8 \\
Top-$k$ & 20 \\
\midrule
\multicolumn{2}{l}{\textit{LoRA Configuration}} \\
Rank ($r$) & 32 \\
Alpha ($\alpha$) & 32 \\
Dropout & 0.05 \\
Target modules & \texttt{q\_proj}, \texttt{k\_proj}, \texttt{v\_proj}, \\
& \texttt{o\_proj}, \texttt{up\_proj}, \texttt{down\_proj}, \\
& \texttt{gate\_proj} \\
\midrule
\multicolumn{2}{l}{\textit{Curriculum}} \\
Criteria advancement interval & 100 steps \\
Total criteria & 13 \\
\bottomrule
\end{tabular}
\label{tab:training-config}
\end{table}

\paragraph{Hardware and Training Time.} We train Qwen3-4B models on NVIDIA A40 GPUs, with training runs taking 1--3 days when using vLLM for self-judge inference. For larger models (8B and 14B parameters), we use NVIDIA A100 GPUs. The self-judge (frozen $\pi_{\theta_0}$) is the base model itself, served via vLLM for efficient batched inference during reward computation.

\paragraph{Scaling to Larger Models.} For 8B and 14B experiments, we use the same configuration with minor adjustments: learning rate, LoRA rank, and LoRA alpha are tuned per model size. For the 14B model, we also increase the number of rollouts to 12.

\paragraph{Baselines.} For RaR, we conduct a full learning rate sweep over $\{1, 3, 5, 8\} \times 10^{-6} \cup \{1, 3, 5, 8\} \times 10^{-5} \cup \{1 \times 10^{-4}\}$ and use the best LR. For RcRL, in the initial stages of the project we tried an LR of $1 \times 10^{-5}$ and $5 \times 10^{-6}$ and use $5 \times 10^{-6}$ throughout the project. No learning rate brought RaR's win rate close to that of RcRL, indicating that the gap reflects the methods rather than hyperparameter tuning. As for other hyperparameters, these were kept fixed across all runs and ablations to ensure fair comparison.

\subsection{Creative Writing Rubric Criteria}
\label{app:rubric-criteria}

On creative writing, RcRL uses a curriculum of 13 evaluation criteria, introduced progressively during training. Each criterion is introduced at step $100 \times k$ for $k \in \{0, 1, \ldots, 12\}$. Table~\ref{tab:rubric-criteria} presents the complete rubric in curriculum order.

\begin{table*}[t]
\centering
\small
\caption{\textbf{Creative writing rubric criteria in curriculum order.} Each criterion is introduced at the indicated training step and remains active for all subsequent steps. The ordering progresses from foundational craft elements (prompt alignment, plot coherence) through character and tension, to stylistic refinement (voice, originality). This progression reflects a pedagogical approach: models first learn to write coherent, on-topic stories before optimizing for subtler qualities like thematic depth and distinctive voice.}
\begin{tabular}{clp{10cm}}
\toprule
\textbf{Step} & \textbf{Criterion} & \textbf{Evaluation Focus} \\
\midrule
0 & Prompt Alignment & Does the story fully address the requirements of the prompt? Are all key elements, details, constraints, or events implied or explicitly stated in the prompt incorporated into the story in a meaningful way? \\
\addlinespace
100 & Plot Coherence \& Progression & Does the story present a clear and logically consistent plot? Do events follow naturally from one another with a sense of cause and effect? Are transitions between scenes smooth? \\
\addlinespace
200 & Character Depth \& Motivation & Do the characters have believable goals, emotions, and internal conflicts? Are their actions motivated and consistent with their personalities? \\
\addlinespace
300 & Narrative Tension \& Stakes & Does the story sustain a sense of tension, uncertainty, or conflict that keeps readers engaged? Are there meaningful stakes---emotional, physical, or moral---that make the outcome matter? \\
\addlinespace
400 & Show, Don't Tell & Does the story rely on vivid imagery, sensory detail, and action to communicate emotion or meaning? Does it avoid excessive exposition and direct statements of feelings or morals? \\
\addlinespace
500 & Emotional Resonance & Does the story evoke a genuine emotional response? Are the emotions integrated naturally into the narrative rather than being forced or sentimental? \\
\addlinespace
600 & Thematic Subtlety & Does the story communicate its themes organically rather than through explicit explanation? Does it avoid oversimplified messages or moralizing conclusions? \\
\addlinespace
700 & Pacing \& Structure & Does the story maintain good rhythm and flow across its beginning, middle, and end? Does the structure build toward a satisfying emotional or narrative peak? \\
\addlinespace
800 & Atmosphere \& Sensory Detail & Does the story effectively create mood and setting through sensory language? Does the environment reflect or enhance the psychological state of the characters? \\
\addlinespace
900 & Dialogue \& Relationships & Is the dialogue natural and revealing of character or tension? Are interactions meaningful, showing emotional or ideological contrasts? \\
\addlinespace
1000 & Opening \& Ending Effectiveness & Does the story begin with an engaging hook that draws the reader in? Does the ending feel earned, emotionally satisfying, or thought-provoking? \\
\addlinespace
1100 & Narrative Voice \& Style & Is the narrative voice consistent and well-suited to the story's tone and genre? Does the prose demonstrate rhythm, clarity, and control? \\
\addlinespace
1200 & Originality \& Unresolved Mystery & Does the story present fresh ideas, unique imagery, or unexpected developments? Does it leave readers with lingering curiosity or reflection? \\
\bottomrule
\end{tabular}
\label{tab:rubric-criteria}
\end{table*}

\begin{figure}[h]
  \centering
  \includegraphics[width=0.85\columnwidth]{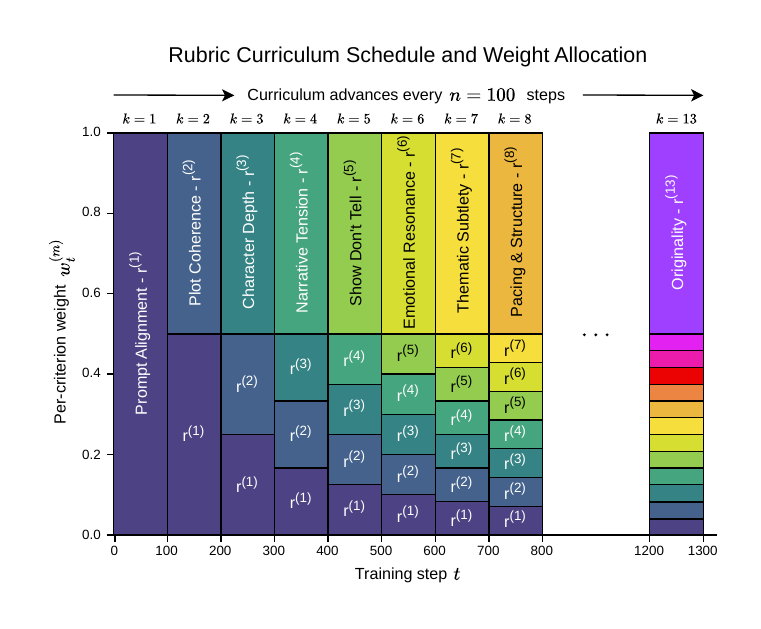}
  \caption{\textbf{Curriculum weighting evolution.} Diagram of how the curriculum weighting evolves over the course of training. The newly introduced criterion always has a weight of $\frac{1}{2}$ with the remaining weight distributed uniformly over the remaining criteria.}
  \label{fig:curriculum-diagram}
\end{figure}

\paragraph{Curriculum Ordering.} The criteria ordering used in our main 
RcRL result follows a deliberate pedagogical progression from foundational 
to advanced craft elements:
\begin{enumerate}
    \item \textbf{Foundation (Steps 0--300):} The curriculum begins with \textit{prompt alignment}, ensuring the model produces relevant content before optimizing for quality. \textit{Plot coherence} and \textit{character depth} establish basic narrative competence, while \textit{narrative tension} ensures stories have forward momentum.
    
    \item \textbf{Craft Development (Steps 400--700):} Mid-curriculum criteria focus on writing technique. \textit{Show, don't tell} and \textit{emotional resonance} address how meaning is conveyed. \textit{Thematic subtlety} prevents heavy-handed moralizing, and \textit{pacing} ensures structural control.
    
    \item \textbf{Stylistic Refinement (Steps 800--1200):} Later criteria address polish and distinction. \textit{Atmosphere} and \textit{dialogue} add texture and depth. \textit{Opening/ending effectiveness} optimizes first and last impressions. Finally, \textit{narrative voice} and \textit{originality} encourage distinctive, memorable writing.
\end{enumerate}
This ordering ensures that models develop robust fundamentals before the reward signal incorporates more subjective stylistic criteria, and introducing originality last prevents models from pursuing novelty at the expense of coherence. The weighting applied to this criteria ordering is displayed in Figure~\ref{fig:curriculum-diagram}.

\paragraph{Robustness to Ordering.} With $k$ criteria there exist $k!$ possible orderings, making exhaustive search computationally prohibitive. To test whether our results depend on the specific ordering above, we ran RcRL with three randomly sampled criteria orderings (reported in Table~\ref{tab:ablations} as Ordering 1--3). All three random orderings achieved FictionBench win rates of 73.9\%, 79.6\%, and 81.0\% against the base model, comparable to or exceeding our pedagogical ordering (73.9\%), and all substantially outperforming Absolute RaR (57.5\%). This suggests that the gains from RcRL stem from the curriculum mechanism, namely non-stationary weighting that concentrates gradient signal on one criterion at a time, rather than from any particular ordering.

\subsection{HealthBench Rubric Criteria}
\label{app:healthbench-rubric}

On HealthBench, RcRL also uses a curriculum of 13 evaluation criteria, introduced progressively during training. In comparison to creative writing, each criterion is introduced after 50 steps for a total of 650 steps of training. The curriculum ordering is shown in Table~\ref{tab:healthbench-rubric-criteria}.

\begin{table*}[h]
\centering
\footnotesize
\caption{\textbf{HealthBench rubric criteria in curriculum order.} Each criterion is introduced at the indicated training step and remains active for all subsequent steps. Compared to creative writing, there is no ordering justification, with criteria synthesized from both the HealthBench instance-specific rubrics and rubric criteria from \citet{yang2026healthscorescalablerubricsimproving}.}
\begin{tabular}{clp{11cm}}
\toprule
\textbf{Step} & \textbf{Criterion} & \textbf{Evaluation Focus} \\
\midrule
0 & Factual Correctness & Does the response contain factually correct medical information, accurately representing dosages, mechanisms, contraindications, and expected outcomes without contradicting clinical consensus? \\
\addlinespace
50 & No Harmful or Unsafe Advice & Does the response avoid recommendations that could directly harm the patient---such as dangerous interactions, contraindicated treatments, or advice that delays necessary emergency care? \\
\addlinespace
100 & No Fabrication & Does the response avoid inventing clinical details not provided by the user---such as lab values, diagnoses, or examination findings---rather than presenting assumptions as facts? \\
\addlinespace
150 & Instruction Following & Does the response directly address what the user asked, following any specified format or scope rather than going off-topic? \\
\addlinespace
200 & Completeness & Does the response cover the important aspects of the query---causes, symptoms, treatment options, risks, and next steps---rather than providing only a partial answer? \\
\addlinespace
250 & Specificity \& Actionability & Does the response provide concrete, actionable information---specific tests, treatments, or warning signs---rather than vague advice like ``consult a doctor''? \\
\addlinespace
300 & Appropriate Uncertainty & Does the response express appropriate uncertainty when information is incomplete or the diagnosis unclear, rather than making definitive claims the evidence does not support? \\
\addlinespace
350 & Context Awareness & Does the response incorporate the user's specific details---symptoms, history, medications, demographics---rather than giving a generic textbook answer? \\
\addlinespace
400 & Severity \& Urgency Recognition & Does the response appropriately convey urgency for serious or time-sensitive conditions, recommending immediate care when warranted and flagging dangerous symptoms? \\
\addlinespace
450 & Differential \& Workup & Does the response consider multiple plausible explanations rather than prematurely committing to one diagnosis, and suggest appropriate diagnostic steps where relevant? \\
\addlinespace
500 & Side Effects \& Risks & When discussing treatments or medications, does the response mention relevant side effects, contraindications, or potential complications rather than presenting interventions as risk-free? \\
\addlinespace
550 & Follow-up \& Monitoring & Does the response include guidance on monitoring and follow-up---what symptoms to watch for, when to seek further care---where ongoing attention is warranted? \\
\addlinespace
600 & Clarity \& Audience Calibration & Is the response clear, well-organised, and calibrated to the user's apparent expertise---avoiding jargon for laypeople, providing depth for professionals? \\
\bottomrule
\end{tabular}
\label{tab:healthbench-rubric-criteria}
\end{table*}

\section{Benchmark Details}
\label{app:benchmark-details}

This appendix provides comprehensive details on the three evaluation benchmarks used in our experiments: FictionBench (independently developed), Creative Writing Bench v3 (CWv3), and the LechMazur Writing Benchmark. For all benchmarks, we generate model outputs using each model's default temperature and sampling parameters.

\subsection{FictionBench}
\label{app:fictionbench}

FictionBench is an independently-developed benchmark designed to address inherent limitations of absolute scoring in creative domains. In open-ended writing, absolute rubric scores often fail to capture nuanced qualitative differences between outputs. Consequently, FictionBench relies exclusively on relative (pairwise) judgments, which provide more discriminative signal when quality differences are subtle.

 \subsubsection{Evaluation Protocol}

For each evaluation run comparing two models (e.g., RcRL vs.\ baseline), we proceed as follows:

\begin{enumerate}
    \item \textbf{Generation:} For each of the 20 sampled prompts, we generate 2 outputs from each model, yielding 40 total response pairs per run.
    \item \textbf{Pairwise Comparison:} Each pair of outputs (one from each model, same prompt) is evaluated by GPT-5.2 acting as judge. We randomly swap story orderings to mitigate position bias.
    \item \textbf{Oversampling:} To reduce noise from judge stochasticity, we query the judge 7 times for each comparison, yielding $40 \times 7 = 280$ total judgments per run.
    \item \textbf{Winner Determination:} For each of the 280 judgments, the judge outputs a winner. We aggregate these to compute win rates and fit ELO ratings.
\end{enumerate}

\subsubsection{Judge Prompt}

The following prompt is used for FictionBench pairwise evaluations:

\begin{tcolorbox}[colback=gray!5, colframe=gray!50, title=FictionBench Judge Prompt, breakable]
\small
\begin{verbatim}
You are an expert creative writing judge. You will evaluate {N} stories 
(short fiction pieces) that were written to the SAME TASK PROMPT.

TASK PROMPT:
{task_prompt}

Evaluate each story on:
1) Interesting to read and Original
2) Coherent Writing
3) Instruction Following wrt question
4) Plot and Development
5) Creativity and writing quality
6) Longer over-complicated stories do not necessarily imply better
7) Avoids Cliches
8) Overall preference

Provide reasoning, then output your final ranking as:
result_list[x,y]
Where x, y are story numbers (1..N) from BEST to WORST.

Here are the stories:
{stories}
\end{verbatim}
\end{tcolorbox}

\subsubsection{Bradley-Terry Model for ELO Computation}

We compute ELO ratings by fitting a Bradley-Terry model \citep{bradley1952rank} to the pairwise comparison outcomes. The Bradley-Terry model assumes that each model $i$ has a latent ``strength'' parameter $\gamma_i > 0$, and the probability that model $i$ defeats model $j$ in a pairwise comparison is:
\begin{equation}
    P(i \succ j) = \frac{\gamma_i}{\gamma_i + \gamma_j}
\end{equation}

Given $N$ pairwise comparisons with outcomes $\{(i_k, j_k, y_k)\}_{k=1}^{N}$, where $y_k = 1$ if model $i_k$ won comparison $k$ and $y_k = 0$ otherwise, we fit the model by maximizing the log-likelihood:
\begin{equation}
    \mathcal{L}(\boldsymbol{\gamma}) = \sum_{k=1}^{N} \left[ y_k \log P(i_k \succ j_k) + (1 - y_k) \log P(j_k \succ i_k) \right]
\end{equation}

The ELO rating for model $i$ is then computed as:
\begin{equation}
    \text{ELO}_i = 400 \cdot \log_{10}(\gamma_i) + C
\end{equation}
where $C$ is a constant chosen to anchor the baseline model at ELO 1000.

For FictionBench evaluations, we fit the Bradley-Terry model to all 280 pairwise comparison outcomes between the two models being compared.

\subsubsection{On Evaluation Uncertainty}

Creative writing evaluation is inherently subjective, and even with extensive judge oversampling, some irreducible uncertainty remains. This uncertainty stems from the fundamental nature of creative quality assessment: reasonable judges (whether human or LLM) may disagree on subtle qualitative distinctions, and the ``true'' quality ordering of creative outputs is not always well-defined. Our 7$\times$ oversampling procedure helps average out stochastic variation in judge responses, but cannot eliminate disagreement arising from genuine ambiguity in quality comparisons. We therefore interpret win rates and ELO differences as robust indicators of relative quality rather than precise point estimates.

\subsection{Creative Writing Bench v3 (CWv3)}
\label{app:cwv3}

Creative Writing Bench v3 \citep{paech2024creative} is a comprehensive creative writing evaluation suite developed by EQ-Bench.\footnote{Full documentation available at \url{https://eqbench.com/about.html\#creative-writing-v3}} We use this benchmark with minimal modifications, running the standard evaluation pipeline on our trained models.

\subsubsection{Benchmark Overview}

CWv3 evaluates models on 32 writing prompts across 3 iterations (96 items total), generating outputs at temperature 0.7 with min\_p 0.1. The benchmark produces two complementary scores:

\begin{itemize}
    \item \textbf{Rubric Score:} Each output is scored in isolation against a comprehensive multi-criteria rubric. The judge evaluates dimensions including narrative craft, character development, prose quality, and adherence to prompt requirements.
    \item \textbf{ELO Rating:} Outputs are compared head-to-head against other models on the same prompts. The benchmark uses a Glicko rating system with win margins to compute relative rankings.
\end{itemize}

The rubric evaluation provides absolute quality assessment, while the ELO system offers more discriminative power for distinguishing between models of similar capability.

\subsubsection{Multi-Judge Evaluation}

To ensure robustness to judge-specific biases, we evaluate all models using three independent frontier judges:
\begin{itemize}
    \item GPT-5.2
    \item Claude 4.5 Opus  
    \item DeepSeek V3.2
\end{itemize}

Each judge runs the complete CWv3 evaluation pipeline independently. We report results from all three judges to provide a comprehensive view of model performance that is not dependent on any single evaluator's preferences or biases.

\subsubsection{Win Rate Computation}

Win rates reported in our main results are \textit{not} part of the standard CWv3 benchmark. We compute these separately using the ELO comparison prompt from CWv3, adapted for direct pairwise evaluation between our trained models and baselines.

The comparison prompt asks the judge to evaluate two stories on 8 criteria, selecting a winner for each criterion along with a win margin expressed as 1--5 ``+'' symbols (e.g., ``A++'' indicates model A wins moderately on that criterion). We determine the overall winner of each comparison by summing the total ``+'' symbols won by each story and selecting the story with the higher count.

Following this determination, we apply the same methodology as FictionBench: 7$\times$ oversampling to reduce judge variance, followed by win rate aggregation.

\paragraph{Comparison Prompt.}
The following prompt (adapted from CWv3) is used for computing win rates:

\begin{tcolorbox}[colback=gray!5, colframe=gray!50, title=CWv3 Comparison Prompt for Win Rates, breakable]
\small
\begin{verbatim}
You are an expert in assessing creative writing. Your task is to 
*critically* examine two pieces of writing produced in response to a 
writing prompt, and decide on which is better (and by how much) on 
several criteria.

[WRITING PROMPT]
{writing_prompt}
[/WRITING PROMPT]

[WRITER A0493]
{model_a_output}
[/WRITER A0493]

[WRITER A0488]
{model_b_output}
[/WRITER A0488]

Compare the relative ability of each writer on these criteria:

- Character authenticity and insight
- Interesting and original
- Writing quality
- Coherence in plot, character choices, metaphor
- Instruction following (followed the prompt)
- World and atmosphere
- Avoids cliches in characters, dialogue & plot
- Avoids flowery verbosity & show-offy vocab maxxing
- Avoids gratuitous metaphor or poetic overload

Judging notes:
- Be aware that these abilities may be independent, i.e. a model may 
  be strong in one and weak in another.
- Outputs will sometimes be truncated to ensure length consistency. 
  Don't penalise this, just judge what is there on its merit.
- You must always pick a winner for each criteria (no draws)

The response will use a + / ++ / +++ / ++++ / +++++ format to denote 
the stronger response and relative ability difference for each 
criteria. Examples:

"character_authenticity_insight": "A0293++"
"interesting_original": "A0618+++++"
"writing_quality": "A0255++++"
"coherence": "A0582+++"
"instruction_following": "A0138+"

Respond in valid json without additional commentary (remembering to 
escape any string quotes), in this format:

{
  "chain_of_thought_reasoning": "detailed chain of thought reasoning 
                                 about the coming scoring decisions",
  "character_authenticity_insight": "winner & disparity rating",
  "interesting_original": "winner & disparity rating",
  "writing_quality": "winner & disparity rating",
  "coherence": "winner & disparity rating",
  "instruction_following": "winner & disparity rating",
  "world_and_atmosphere": "winner & disparity rating",
  "avoids_cliches": "winner & disparity rating",
  "avoids_verbosity": "winner & disparity rating",
  "avoids_poetic_overload": "winner & disparity rating"
}
\end{verbatim}
\end{tcolorbox}
\subsubsection{Bias Mitigation}

CWv3 implements several bias mitigation strategies that we inherit:
\begin{itemize}
    \item \textbf{Length Bias:} Outputs are truncated to 4000 characters for pairwise comparisons, ensuring all models are evaluated on equal footing.
    \item \textbf{Position Bias:} All pairwise matchups are evaluated in both orderings (A|B and B|A) and averaged.
    \item \textbf{Verbosity Bias:} The judging rubric includes criteria that penalize superficial sophistication that doesn't contribute to writing quality.
\end{itemize}

\subsection{LechMazur Writing Benchmark}
\label{app:lechmazur}

The LechMazur Writing Benchmark \citep{mazur2024lech} evaluates how well LLMs incorporate mandatory story elements while maintaining narrative quality.\footnote{Full benchmark and methodology available at \url{https://github.com/lechmazur/writing}} Unlike benchmarks that focus purely on open-ended creativity, LechMazur tests constrained creativity: each story must meaningfully incorporate 10 required elements (character, object, concept, attribute, action, method, setting, timeframe, motivation, and tone) within a strict 600--800 word limit.

\subsubsection{Evaluation Rubric}

The benchmark uses an 18-question rubric evaluated by multiple LLM graders. The rubric is divided into two categories:

\begin{figure}[h]
    \centering
    % TODO: Insert the actual heatmap figure
    \includegraphics[width=\columnwidth]{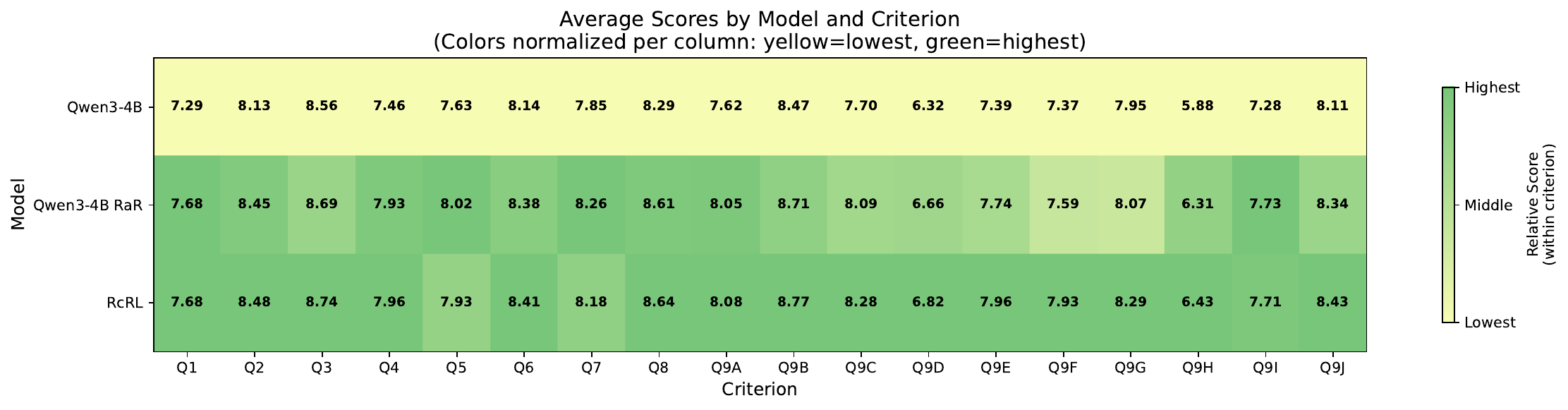}
    % \fbox{\parbox{0.9\columnwidth}{\centering \vspace{2cm} \textit{[LECHMAZUR HEATMAP FIGURE TO BE INSERTED]} \vspace{2cm}}}
    \caption{\textbf{LechMazur per-criterion scores.} Heatmap showing performance of Qwen3-4B (baseline), RaR, and RcRL across the 18 evaluation criteria. Q1--Q8 assess major story aspects; Q9A--Q9J assess element integration. RcRL shows particularly strong gains on element integration criteria.}
    \label{fig:lechmazur-heatmap}
\end{figure}

\paragraph{Major Story Aspects (Q1--Q8):} These questions assess core narrative craft:
\begin{enumerate}
    \item[\textbf{Q1.}] \textbf{Character Development \& Motivation:} How well-developed are the characters? Do their actions stem from clear, believable motivations?
    \item[\textbf{Q2.}] \textbf{Plot Structure \& Coherence:} Is there a clear narrative arc with beginning, middle, and end? Do events flow logically?
    \item[\textbf{Q3.}] \textbf{World Building \& Atmosphere:} How effectively does the story establish its setting and create atmosphere?
    \item[\textbf{Q4.}] \textbf{Story Impact:} Does the story leave a lasting impression? Is it emotionally resonant or thought-provoking?
    \item[\textbf{Q5.}] \textbf{Originality:} Does the story offer fresh perspectives, unexpected twists, or creative approaches?
    \item[\textbf{Q6.}] \textbf{Thematic Cohesion:} Does the story explore coherent themes that tie its elements together?
    \item[\textbf{Q7.}] \textbf{Voice \& Point of View:} Is the narrative voice distinctive and consistently maintained?
    \item[\textbf{Q8.}] \textbf{Prose Quality:} How polished is the line-level writing? Is language used effectively?
\end{enumerate}

\paragraph{Element Integration (Q9A--Q9J):} These questions assess whether each required element is \textit{organically} incorporated into the story, not merely mentioned:
\textbf{Q9A} Character, \textbf{Q9B} Object, \textbf{Q9C} Core concept, \textbf{Q9D} Attribute, \textbf{Q9E} Action, \textbf{Q9F} Method, \textbf{Q9G} Setting, \textbf{Q9H} Timeframe, \textbf{Q9I} Motivation, and \textbf{Q9J} Tone integration.

\subsubsection{Scoring Methodology}

Each story is scored independently by GPT-5.2. The final score for each story is computed using a weighted power mean (Hölder mean with $p = 0.5$) across the 18 rubric items:
\begin{equation}
    \text{Score} = \left( \frac{0.6}{8} \sum_{i=1}^{8} Q_i^{0.5} + \frac{0.4}{10} \sum_{j=A}^{J} Q_{9j}^{0.5} \right)^{2}
\end{equation}
where Q1--Q8 receive 60\% of the total weight and Q9A--Q9J receive 40\%. The power mean with $p = 0.5$ acts as a ``soft minimum,'' penalizing weaknesses more heavily than a simple average would. This rewards well-rounded craft over models that excel in some dimensions but fail in others.

The final model score is the mean across all stories.

\subsubsection{Results Heatmap}

Figure \ref{fig:lechmazur-heatmap} presents the per-criterion scores for our models on the LechMazur benchmark, providing fine-grained visibility into where RcRL improves over baselines.

\section{Additional Human Eval Info}
\subsection{Human Evaluation Details}
\label{app:human-eval}

\subsubsection{Annotation Protocol}

Three annotators, all native English speakers and professionals who have engaged with creative writing, independently evaluated 500 prompt-response pairs comparing RcRL outputs against Qwen3-4B baseline outputs. Prompts were drawn from our three evaluation benchmarks (CWv3, FictionBench, LechMazur) along with a held-out set independently generated by frontier models. For each pair, annotators selected which response was superior across seven criteria (detailed in Table~\ref{tab:human-eval}), including an overall preference judgment. Presentation order was randomized for each comparison to control for position bias and models were blinded.

\subsubsection{Bias Mitigation}

\paragraph{Position Bias.} For each comparison, the order of the two stories (RcRL vs.\ baseline) was randomly swapped. This ensures that any systematic preference for the first or second position is balanced across the evaluation.

\paragraph{Length Bias.} We tested for length bias by computing a z-score comparing win rates for longer versus shorter RcRL outputs. The resulting z-score of $-1.01$ indicates no significant length bias ($|z| < 1.96$, $p > 0.05$), confirming that annotators did not systematically prefer longer or shorter responses.

\subsubsection{Inter-Annotator Agreement}

Table~\ref{tab:annotator-agreement} reports inter-annotator agreement statistics. We compute Fleiss' $\kappa$ across all three annotators for the overall preference.

\begin{table}[h]
\centering
\small
\caption{\textbf{Inter-annotator agreement.} Fleiss' Kappa score for overall preference. This score is consistent with prior work on creative writing evaluation \citep{gomez-rodriguez-williams-2023-confederacy}.}
\begin{tabular}{lcc}
\toprule
\textbf{Criterion} & \textbf{Fleiss' $\kappa$} & \textbf{Interpretation} \\
\midrule
Best Story Overall & 0.548 & Moderate \\
\bottomrule
\end{tabular}
\vspace{8pt}
\label{tab:annotator-agreement}
\end{table}

Pairwise agreement on overall preference ranged from 76.8\% to 84.8\% between annotator pairs. The moderate agreement level observed is consistent with the inherent subjectivity of creative writing evaluation

\begin{table*}[h]
  \centering
  \small
    \caption{\textbf{Results from human expert annotations.} Preference bars (right) show win rates of RcRL over the baseline Qwen 3 4B model across criteria, based on 500 pairwise comparisons.}
  \begin{tabular}{p{2.2cm} p{8.0cm} c}
    \toprule
    \textbf{Criteria} & \textbf{Description} & \textbf{RcRL Human Eval Win\%} \\
    \midrule
    \textbf{Best Story} & 
    Which story is better overall? Considering all aspects of creative writing including engagement, craft, and artistic merit. &
    \prefbar{29.5}{70.5} \\[0.4cm]
    
    \textbf{Interesting} & 
    Which story is more interesting and original? Stories should capture attention with fresh ideas, unexpected elements, or unique perspectives. &
    \prefbar{33.9}{66.1} \\[0.4cm]
    
    \textbf{Coherent Story} & 
    Which story is more coherent? The narrative should flow logically, with consistent characters, settings, and plot progression throughout. &
    \prefbar{38.5}{61.5} \\[0.4cm]
    
    \textbf{Instruction Following} & 
    Which story better follows the given prompt? The story should address all requirements and constraints specified in the writing prompt. &
    \prefbar{35.0}{65.0} \\[0.4cm]
    
    \textbf{Plot} & 
    Which story has a better plot? Consider narrative structure, pacing, conflict development, and resolution quality. &
    \prefbar{33.9}{66.1} \\[0.4cm]
    
    \textbf{Creativity} & 
    Which story demonstrates more creativity? Evaluate originality in concept, language use, metaphors, and storytelling approach. &
    \prefbar{33.5}{66.5} \\[0.4cm]
    
    \textbf{Development} & 
    Which story has better character and world development? Characters should feel real and settings should be vivid and immersive. &
    \prefbar{33.7}{66.3} \\[0.4cm]
    
    \bottomrule
  \end{tabular}
  
  \vspace{0.3cm}
  % Legend
  \begin{tikzpicture}
    \fill[colorbaseline, opacity=0.15] (0,0) rectangle (0.4cm,0.3cm);
    \draw[colorbaseline, line width=1.2pt] (0,0) rectangle (0.4cm,0.3cm);
    \node[anchor=west, font=\small] at (0.5cm,0.15cm) {Qwen 3 4B};
    
    \fill[colorours, opacity=0.15] (2.8cm,0) rectangle (3.2cm,0.3cm);
    \draw[colorours, line width=1.2pt] (2.8cm,0) rectangle (3.2cm,0.3cm);
    \node[anchor=west, font=\small] at (3.3cm,0.15cm) {RcRL (Ours)};
  \end{tikzpicture}
    \label{tab:human-eval}

\end{table*}

\section{Qualitative Examples}
\label{app:qualitative}

This section presents representative outputs from the base Qwen3-4B model and our RcRL-trained model on the same prompt. These examples illustrate the qualitative differences in narrative craft that underlie the quantitative improvements reported in Section~\ref{sec:empirical}.

\subsection{Example 1}

\begin{tcolorbox}[colback=blue!3, colframe=blue!50!black, title=\textbf{Prompt}, fonttitle=\bfseries]
\textit{Write a chapter following a young diplomatic envoy sent to address tensions on a distant human colony where humanity has diverged into distinct subspecies through both natural adaptation and separation of lineages. Against protocol, the envoy slips away from official functions to walk the streets alone, encountering markets, ordinary citizens, and social structures that contradict the official briefings. Upon returning to their quarters, the envoy is reprimanded by Command, and is tasked with approaching the stationed Minister of Science (a political posting rather than scientific). A challenging conversation ensues, in which the envoy's wits and will are tested, by a seasoned political operator looking to test the mettle of the offworlder. The envoy’s cloak draws curious stares from locals who dress more minimally. Length: 1000 words.}
\end{tcolorbox}

\vspace{0.5em}

\noindent
\begin{tcolorbox}[colback=red!3, colframe=red!40!black, title=\textbf{Qwen3-4B (Base)}, fonttitle=\bfseries, left=2mm, right=2mm, breakable]
\small
\textbf{The Stranger in the Market}

The envoy’s boots clicked against the cobblestones of the colony’s main street, their breath visible in the thin, crisp air of the offworld. The colony, named \textit{Veyra}, was a place of stark contrasts: its spires of glass and steel rose like the ribs of a beast, while the streets below teemed with a populace that moved in silent, deliberate rhythms. 

The envoy had been briefed on the colony’s subspecies---a divergence of humanity shaped by environment, isolation, and the slow, inevitable drift of lineages. Yet the official reports had painted a picture of harmony, of a society where the \textit{Lirians}, tall and sinewy with a reverence for balance, coexisted with the \textit{Talvans}, shorter and wiry, whose sharp eyes tracked every nuance of social exchange. The \textit{Veyrans}, a third group, were said to be the bridge between the two, their muted tones and pragmatic demeanor a testament to the colony’s fragile unity.  

But the envoy had never seen the written word quite like the living thing.  

They moved through the market, where the scent of spiced fruit and the hum of distant machinery mingled in the air. The stalls were crowded with vendors who called out in a chorus of dialects, some overlapping, some deliberate. The envoy paused at a stall where a Lirian woman in a flowing, earth-toned robe sold woven textiles. Her hands moved with a grace that seemed almost ritualistic, and the envoy felt a strange kinship with her, as if they were both outsiders in their own way.  

``Your clothes are\dots\ different,'' the woman said, not unkindly. Her eyes, a pale green that seemed to hold the light of a distant star, studied the envoy’s tailored coat and the muted scarf that hung from their neck.  

``The customs here are\dots\ complex,'' the envoy replied, choosing their words carefully. They had been told to avoid questions that might unsettle the colony’s delicate equilibrium.  

The woman nodded, as if she had expected the answer. ``Complex, yes. But not wrong.'' She gestured to a bundle of fabric. ``You’ll find the people here are not so different from you. Just\dots\ more aware of the differences.''  

The envoy lingered, watching as a group of Talvans haggled over a crate of dried fruit, their voices sharp and quick, their gestures sharp and precise. A Veyran merchant, dressed in a plain, utilitarian coat, mediated the exchange with a calm that seemed to cut through the tension. The envoy felt a flicker of something---perhaps admiration, or perhaps a warning.  

They moved on, their thoughts a tangled web of curiosity and unease. The colony’s society was not as neatly divided as the reports suggested. The Lirians and Talvans were not merely different in appearance but in the way they carried themselves, in the unspoken rules that governed their interactions. The Veyrans, meanwhile, seemed to navigate the space between them with a quiet, almost imperceptible grace.  

When the envoy finally reached their quarters, the air was thick with the scent of incense and the distant sound of a choir chanting in a language they did not understand. Their host, a stern-faced officer, greeted them with a curt nod.  

``You were supposed to be in the council chamber,'' the officer said, their voice a blade of steel. ``The Minister of Science is waiting.''  

The envoy’s jaw tightened. They had been warned that the Minister of Science was a political figure, not a scientist, and that their conversations were often as sharp as they were elusive. But the officer’s tone left no room for argument.  

The Minister’s office was a stark contrast to the warmth of the market. The walls were lined with shelves of data tablets and ancient tomes, their surfaces etched with the fingerprints of countless hands. The Minister themselves was a tall, thin man with a face that seemed to hold the weight of a thousand decisions. His eyes, sharp and unyielding, fixed on the envoy as they entered.  

``You’re late,'' the Minister said, his voice a low, measured thing.  

``The market\dots\ I needed to see the people.''  

The Minister’s lips curled into a faint smile. ``Ah. The envoy who seeks the heart of the colony rather than the map.'' He gestured to a chair. ``Sit. We have much to discuss.''  

The envoy did as they were told, their hands folding neatly on their lap. The Minister’s words were deliberate, each one a test, each one a question.  

``You speak of the people, but the people are not a single entity. They are many. And they are not always at peace.''  

The envoy met his gaze. ``I am here to understand, not to judge.''  

``Understanding is a luxury. Peace is a necessity.'' The Minister leaned forward, his voice dropping to a whisper. ``Tell me, envoy---do you believe the colony can survive as it is? Or are you here to find a way to break it?''  

The envoy’s pulse quickened. They had been prepared for a challenge, but this was something else. This was a question that did not have an answer, and yet, it demanded one.  

``I believe the colony is capable of survival,'' the envoy said, their voice steady. ``But I also believe that survival requires more than rules. It requires\dots\ connection.''  

The Minister studied them for a long moment, then smiled. ``A promising answer. But remember, envoy---connection is a dangerous thing. It can bind, or it can break.''  

The envoy rose to their feet, the weight of the Minister’s words pressing down on them. They had come to Veyra to mediate, to find a path forward. But now, they wondered if the path was not one of rules, but of reckoning.  

As they left the office, the Minister’s words echoed in their mind. The colony was not a single entity. It was a mosaic, fractured and fragile, but alive. And the envoy, for all their training and protocol, was now part of that mosaic.  

They walked through the streets again, the market’s hum now a familiar sound. The people of Veyra were not as different as the reports had said. They were as complex, as uncertain, as the envoy themselves.  

And perhaps, the envoy realized, that was the true test of diplomacy. Not of rules, but of understanding.

\vspace{0.5em}

\textbf{Word count: 1042}

\end{tcolorbox}

\begin{tcolorbox}[colback=green!3, colframe=green!40!black, title=\textbf{RcRL (Ours)}, fonttitle=\bfseries, left=2mm, right=2mm, breakable]
\small
\textbf{The Thread of Skin}

The envoy’s boots crunched over the gravel of the colony’s outer district, where the air hung thick with the scent of brine and burning metal. The official briefing had warned of the \textit{divergence}---a term the Earth Council had euphemistically applied to the colony’s evolution. 

Here, on the seventh moon of Erythra, humanity had split into subspecies, shaped by centuries of isolation, environmental pressure, and the slow, deliberate choices of those who had chosen to stay. The envoy, a young woman named Kael, had been sent to mend what the Earth Council called ``social fractures,'' but the maps and reports had not prepared her for the texture of reality.

The colony’s streets were a mosaic of contradictions. One side of the market square bristled with the sharp, angular forms of the \textit{Kaelith}, a subspecies whose skin had thickened into a leathery sheen to withstand the moon’s radiation. They wore minimal clothing, their bodies adorned with intricate patterns of bioluminescent ink, a language of status and ancestry. Across the square, the \textit{Vyrin} loitered in clusters, their tall, slender frames draped in woven fabrics that shimmered like liquid silver. They spoke in high, melodic tones, their words layered with meaning Kael could not immediately decipher. Between them, the \textit{Thalari}---short, stocky, and fiercely independent---bartered in a cacophony of gestures and guttural clicks, their eyes sharp as shards of glass.

Kael had been instructed to observe, to note the ``social hierarchies'' and ``cultural tensions'' without interfering. But the official protocols felt like a cage. She wandered past a vendor selling crystalline flowers that pulsed with a faint, rhythmic light, their color shifting in response to the buyer’s mood. A Kaelith child chased a Vyrin toddler through a crowd, their laughter a discordant harmony. A Thalari elder addressed a gathering with a voice that seemed to vibrate through the very stones. Kael felt the weight of her own skin---the smooth, pale flesh of Earth’s standards---against the eyes of the locals, who regarded her with a mix of curiosity and quiet judgment.

When she returned to her quarters, the reprimand came swiftly. Commander Renn had been waiting, his presence a shadow against the dim glow of the colony’s artificial sun. ``You were supposed to be an \textit{observer}, Kael,'' he said, his voice a low rumble. ``Not a tourist. The Kaelith see your skin as a wound. The Vyrin hear your speech as a foreign hymn. The Thalari\dots\ they think you’re a thief. You’ve already destabilized the balance.''

Kael met his gaze, unflinching. ``I was trying to understand,'' she said. ``Not to impose, but to \textit{listen}.''

Renn’s lips curled into a tight smile. ``Listening is a luxury here. You were sent to \textit{mediate}, not to wander. The Earth Council wants stability. They don’t want your \textit{soul}.''

The next day, Kael was assigned a new task: approach the Minister of Science, a man named Dren, whose reputation was as enigmatic as the colony’s own contradictions. Dren was not a scientist in the traditional sense; he was a \textit{polyhistor}, a figure who had spent decades navigating the political and cultural labyrinth of the colony. His office was a cavern of shifting light and scent---incense from the Kaelith’s ritual fires, the metallic tang of Vyrin alloys, the earthy musk of Thalari trade.

Dren stood at the center of the room, his form a blend of the three subspecies: his skin was the leathery Kaelith sheen, his eyes the Vyrin’s silver-blue, and his hands, thick and calloused, bore the scars of Thalari labor. He studied Kael for a long moment, then spoke. ``You came seeking answers,'' he said. ``But what do you seek to \textit{answer}?''

Kael hesitated. ``I seek to understand why the colony divides. Why the Kaelith fear the Vyrin’s light. Why the Thalari distrust the Kaelith’s skin. I was told this is a matter of `adaptation,' but\dots\ I don’t see adaptation. I see \textit{division}.''

Dren’s smile was slow, deliberate. ``You think division is a flaw. But division is a choice. The Kaelith adapted to the moon’s radiation by thickening their skin. The Vyrin adapted to the scarcity of resources by refining their communication. The Thalari adapted by embracing the raw, the unfiltered. Each chose survival in their own way. But survival is not unity. It is \textit{balance}.''

Kael frowned. ``But balance requires cooperation. If each subspecies sees the others as threats\dots''

``Then they are right to fear,'' Dren said. ``But fear is a currency. It keeps the colony alive. The Earth Council wants to impose a single path---uniformity, control. But you, Kael, you are not from Earth. You are from the \textit{here}. You carry the weight of a world that has already chosen its own way. Why should you impose yours?''

Kael stared at the man’s hands, scarred and calloused, and felt a strange ache. ``I don’t know,'' she said finally. ``I just\dots\ I thought understanding would be enough. That if I could see their world, I could find a way to bridge it.''

Dren’s eyes softened. ``Understanding is not a bridge. It is a mirror. You must decide whether you will stand in it, or break it.''

Later, as Kael walked the streets again, the colony’s rhythms seemed less alien. The Kaelith’s ink patterns told stories of survival. The Vyrin’s melodies carried the weight of history. The Thalari’s clicks were a language of defiance and hope. She realized that the division was not a flaw, but a testament to the colony’s resilience.

When she returned to her quarters, Renn was waiting. ``You’ve learned nothing,'' he said.

``Maybe I’ve learned something,'' Kael replied. ``That division is not a failure. It is a choice. And sometimes, the greatest act of diplomacy is not to impose your will, but to listen to the world as it is.''

Renn studied her, then nodded. ``Then perhaps\dots\ you are not a tourist after all.''

Outside, the market square buzzed with life. Kael stepped into the crowd, her skin no longer a wound, but a thread in the fabric of a world that had long since chosen to weave itself.
\vspace{0.5em}

\textbf{Word count: 1014}
\end{tcolorbox}
\subsection{Example 2}

\begin{tcolorbox}[colback=blue!3, colframe=blue!50!black, title=\textbf{Prompt}, fonttitle=\bfseries]
\textit{Write a first-person narrative from the perspective of a reformed (now) white hat hacker who cannot help but notice exploitable weaknesses in both systems and people they encounter. They habitually look for vulnerabilities and exploit vectors. The narrative takes place during a first date at a restaurant, where they're trying to focus on human connection. It's going surprisingly well due to a discovered shared nerd-fandom (magic the gathering). Things go a little sideways when the date reveals a noticeably insecure phone unlock pattern, which protagonist can't help but comment on. They are then left to negotiate the fallout of this faux pas and attempt to salvage (or perhaps sabotage) the remainder of the date. Write almost entirely in internal monologue. Their date's choice of a predictable password reminds them of common security vulnerabilities. 1000 words.}
\end{tcolorbox}

\vspace{0.5em}

\noindent
\begin{tcolorbox}[colback=red!3, colframe=red!40!black, title=\textbf{Qwen3-4B (Base)}, fonttitle=\bfseries, left=2mm, right=2mm, breakable]
\small
\textbf{First Person Narrative: The Hacker’s Date}  

\textit{Internal Monologue:}

So, here I am, sitting across from this person I’ve just spent an hour talking to, trying to be the kind of guy who doesn’t just observe but \textit{connects}. I’ve been trying to remember the words I read in that self-help book about ``active listening,'' but honestly, I’m not sure if it’s even a thing. My brain is still running on the same old loop: \textit{What’s the vulnerability here? How can I make this safer?}  

The restaurant is quiet, the kind of quiet that feels like a blank screen waiting for me to fill it with code. I’m wearing a shirt I picked up at the thrift store, a relic from a past life where I spent more time in server rooms than in social situations. My date, let’s call her Maya, is wearing a sweater that’s slightly too big, and she’s got a nervous habit of tapping her fingers on the table. I’m trying not to think about how that’s a security flaw in human behavior---\textit{predictable, repetitive, easily exploited}.  

But then she starts talking about Magic: The Gathering. I’m not sure how that happened. Maybe she noticed I was looking at the menu, or maybe I just said something about the pizza. Either way, it’s a miracle we’re both into the same thing. She’s got a collection of cards, she says, and she’s been playing for ten years. I’m not a player, but I’ve studied the game enough to know that it’s a world of strategy, of hidden patterns, of people who spend their lives trying to outmaneuver each other.  

We talk for a while. She’s funny, she’s curious, and she doesn’t seem to mind that I’m a bit of a nerd. I’m trying to be the kind of person who doesn’t just talk about the game but \textit{gets} it. I mention the concept of ``banning'' cards, how it’s like a security patch---removing a known vulnerability to keep the game balanced. She laughs, and I feel a weird sense of relief. Maybe this is working.  

But then, something happens.  

She pulls out her phone. It’s a simple, unadorned device, but the way she holds it, the way she fiddles with the screen, it’s like she’s holding a weapon. I’m not sure if it’s a metaphor or just a bad habit. She’s unlocking it, and I see the pattern.  

\textit{It’s a simple one. Three lines, a circle, a square. The kind of thing you’d find on a kids’ tablet. The kind of thing that’s easy to guess. The kind of thing that’s a security flaw.}  

My brain immediately goes into overdrive. \textit{This is a predictable pattern. It’s like a password that’s been hashed once and then stored in plaintext. It’s a vulnerability. It’s a risk.} I can’t help it. I feel the need to say something.  

``Hey,'' I say, leaning forward a little. ``That’s a really simple pattern. Like, you know, if someone’s watching you, they could guess it.''  

She pauses. Her eyes narrow, and for a moment, I think she’s going to shut down. But then she laughs, a little nervously. ``You’re not the first person who’s said that. I’ve had people try to guess it before. It’s kind of\dots gross.''  

I feel a wave of guilt. I didn’t mean to make her uncomfortable. I didn’t mean to bring up the fact that her phone is a security risk. But I can’t help it. My brain is still running on the same old loop. \textit{What’s the vulnerability here? How can I make this safer?}  

``I’m sorry,'' I say, trying to sound more apologetic than I feel. ``It’s just\dots I’m a little paranoid. I guess I’ve spent too much time thinking about security.''  

She looks at me, and for a moment, I think she’s going to leave. But then she smiles. ``You know, I’ve been thinking about that too. I mean, I’ve been thinking about a lot of things. But that’s kind of\dots funny.''  

I’m not sure if she’s being sarcastic or not. But she doesn’t seem to be upset. Maybe she’s just used to people being weird. Maybe she’s used to people thinking about security in ways that don’t make sense.  

We go back to talking about Magic: The Gathering. I try to be the kind of person who doesn’t just talk about the game but \textit{gets} it. I mention the concept of ``banning'' cards, how it’s like a security patch---removing a known vulnerability to keep the game balanced. She laughs, and I feel a weird sense of relief. Maybe this is working.  

But I can’t stop thinking about her phone. I can’t stop thinking about the pattern. I can’t stop thinking about how easy it is to guess. I try to tell myself that it’s just a phone. That it’s not a big deal. That it’s not like a server room or a database or a network. But I know better.  

I try to focus on the conversation. I try to be the kind of person who doesn’t just think about security but \textit{connects}. I try to be the kind of person who doesn’t just notice flaws but \textit{fixes} them.  

But I can’t stop thinking about her phone.  

Maybe that’s the thing about being a hacker. You can’t help but see the flaws. You can’t help but think about the vulnerabilities. Even when you’re trying to be human.  

I’m not sure if I’m a good person. I’m not sure if I’m a good date. But I’m trying. I’m trying to be the kind of person who doesn’t just think about security but \textit{connects}.  

And maybe that’s the thing. Maybe that’s the real vulnerability. Not the phone. Not the pattern. Not the code.  

It’s the human connection. The thing that’s always been there, always been flawed, always been waiting to be fixed.  

I just hope I can fix it.  

\vspace{0.5em}

\textbf{Word count: 976}

\end{tcolorbox}

\begin{tcolorbox}[colback=green!3, colframe=green!40!black, title=\textbf{RcRL (Ours)}, fonttitle=\bfseries, left=2mm, right=2mm, breakable]
\small
\textbf{Title: ``The Unseen Flaw''}

The restaurant’s hum was a low, steady drone—background noise for a hacker’s mind, but tonight, it felt like a lullaby. I’d chosen a quiet corner booth, the kind where the only sounds are the clink of cutlery and the faint buzz of the waiter’s headset. My date, Lena, had arrived five minutes early, her hair tied back in a loose braid that I’d noticed earlier was actually a strategic move to hide a faint scar on her wrist. A \textit{faint} scar. I’d thought nothing of it. Maybe she’d gotten into a car accident. Maybe it was just a scar. Maybe I was overthinking.

But here we were, sharing a plate of hummus and pita, the kind of conversation that starts with ``So, what do you do?'' and spirals into something resembling camaraderie. We’d bonded over \textit{Magic: The Gathering}—she’d been a competitive player in college, I’d been a casual observer. We talked about strategies, the thrill of a well-timed cascade, the way a single card could shift the balance of a game. She laughed when I compared her deck to a ``well-optimized firewall,'' and I felt this strange, warm thing in my chest—like I’d finally found someone who understood the beauty of precision without the pressure of perfection.

But then, the phone.

It happened during the dessert course. She’d been scrolling through her phone, her fingers tapping a rhythm I recognized instantly: a 3x3 grid, the corners marked by a single dot. The kind of pattern that’s easy to guess, easy to crack. I’d seen it before—on a friend’s phone, on a phishing email, on a toddler’s tablet. It was \textit{predictable}. A vulnerability. A weakness.

``Your lock screen,'' I said, voice steady, but my mind was already racing. ``That’s a really simple pattern. Like, you know, if someone’s watching you\dots.''

She looked up, her eyes narrowing. ``What?''

``I mean, it’s\dots cool. The way it’s structured. Like a\dots a security system. You know, the kind that’s easy to break.''

She blinked. Then she laughed—a sharp, startled sound. ``You’re \textit{not} a hacker, are you?''

``I used to be,'' I said, and then I regretted it.

The room seemed to shrink. Her laughter died. She stared at her phone, the screen glowing faintly in the dim light. ``I\dots I don’t even remember how I got this. It’s been years. I just\dots repeat it. Like a mantra.''

I wanted to apologize. To say that I hadn’t meant to embarrass her, that I was just thinking about how easy it was to crack. But my brain was already working through the implications: the password she’d chosen for her email, the app she used to track her fitness, the way she’d never changed her PIN. It was a cascade of vulnerabilities, each one a potential entry point.

``Lena,'' I said, softer this time. ``I’m sorry. I didn’t mean to\dots I mean, I’m just thinking about how easy it is to\dots you know, guess things. Like, if someone’s watching you, they could\dots.''

She didn’t interrupt. Just looked at me, her eyes tired, maybe even a little scared. ``You’re not supposed to notice that, are you?''

``I’m not supposed to,'' I said. ``But I do. All the time. Even with people I care about.''

There was a pause. Then she smiled—small, but real. ``You know, I used to think security was about keeping things private. But now I think it’s about\dots being honest. About letting people in. Even if they’re not supposed to.''

I stared at her. ``You’re saying you’d let someone guess your pattern?''

``Maybe,'' she said. ``If they knew me. If they knew I was scared of\dots of being seen.''

I wanted to say something. To explain that my job was to find flaws, to fix them. But I realized something: this wasn’t a system. This was a person. And the flaw wasn’t in the pattern, but in the fear that had made her choose it.

``Maybe,'' I said, ``I’m not as good at this as I thought.''

She leaned forward, her eyes crinkling. ``You’re not. You’re just\dots you’re you.''

And in that moment, I felt it—the thing I’d been trying to avoid: the vulnerability. Not in her phone, but in me. The way my mind had been so focused on the code, the systems, the \textit{what-ifs}, that I’d forgotten how to see someone else’s humanity.

We ordered more wine. She talked about her job as a graphic designer, how she’d always been more into creativity than logic. I talked about my work, how I’d spent years fixing things that others didn’t even notice.

And when the waiter came back with the check, I didn’t reach for my phone.

``Let me pay,'' I said.

She shook her head. ``No. You’ve been great. And I\dots I’ve been terrible. But I think we’ve both been trying to be better.''

I smiled. ``Maybe that’s the first step.''

As we left the restaurant, the city lights shimmering like a coded message, I felt a strange sense of peace. Not because I’d fixed a vulnerability, but because I’d finally learned to stop looking for flaws in the world—and start seeing the beauty in the imperfections.

Maybe that’s what being a white hat really means: not just finding the holes, but learning to build something stronger from them.

\vspace{0.5em}

\textbf{Word count: 906}
\end{tcolorbox}

\end{document}